\documentclass{article}

\usepackage{microtype}
\usepackage{graphicx}
\usepackage{subfigure}
\usepackage{booktabs} % for professional tables

\usepackage{hyperref}

\usepackage[accepted]{icml2023}

\usepackage{amsthm}
\usepackage{cite}
\usepackage{amsmath,amssymb,amsfonts}
\usepackage{algorithmic}
\usepackage{tabularx}

\usepackage{amsmath}
\usepackage{amssymb}
\usepackage{amsmath,bm}
\usepackage{color}
\usepackage{xcolor,soul}
\usepackage{algorithm}
\usepackage{multirow}
\usepackage{xspace}
\usepackage{soul}

\usepackage[capitalize,noabbrev]{cleveref}

\theoremstyle{plain}

\theoremstyle{definition}

\theoremstyle{remark}

\usepackage[textsize=tiny]{todonotes}

\newdimen\abovecrulesep
\newdimen\belowcrulesep
\makeatletter
\definecolor{demphcolor}{RGB}{144, 144, 144}
\definecolor{mygray}{gray}{0.4}
\definecolor{lightgray}{rgb}{0.9, 0.9, 0.9}

\newlength\savewidth
\newcommand\shline{\noalign{\global\savewidth\arrayrulewidth\global\arrayrulewidth 1pt}\hline\noalign{\global\arrayrulewidth\savewidth}}

\newcommand{\tablestyle}[2]{\setlength{\tabcolsep}{#1}\renewcommand{\arraystretch}{#2}\centering\small}
\makeatletter\renewcommand\paragraph{\@startsection{paragraph}{4}{\z@}{.5em\@plus1ex\@minus.2ex}{-.5em}{\normalfont\normalsize\bfseries}}
\makeatother

\icmltitlerunning{mPLUG-2: A Modularized Multi-modal Foundation Model Across Text, Image and Video}
\newcommand{\modelname}{mPLUG-2 }
\newcommand{\modelnamebase}{mPLUG-2$_\text{Base}$ }
\newcommand{\modelnamedeberta}{mPLUG-2$_\text{Deberta}$ }
\begin{document}

\twocolumn[
\icmltitle{mPLUG-2: A Modularized Multi-modal Foundation Model \\ Across Text, Image and Video}

% It is OKAY to include author information, even for blind
% submissions: the style file will automatically remove it for you
% unless you've provided the [accepted] option to the icml2023
% package.

% List of affiliations: The first argument should be a (short)
% identifier you will use later to specify author affiliations
% Academic affiliations should list Department, University, City, Region, Country
% Industry affiliations should list Company, City, Region, Country

% You can specify symbols, otherwise they are numbered in order.
% Ideally, you should not use this facility. Affiliations will be numbered
% in order of appearance and this is the preferred way.
\icmlsetsymbol{equal}{*}

\begin{icmlauthorlist}
\icmlauthor{Haiyang Xu}{equal,damo}
\icmlauthor{Qinghao Ye}{equal,damo}
\icmlauthor{Ming Yan}{damo}
\icmlauthor{Yaya Shi}{damo}
\icmlauthor{Jiabo Ye}{damo}
\icmlauthor{Yuanhong Xu}{damo}
\icmlauthor{Chenliang Li}{damo}
\icmlauthor{Bin Bi}{damo}
\icmlauthor{Qi Qian}{damo}
\icmlauthor{Wei Wang}{damo}
\icmlauthor{Guohai Xu}{damo}
\icmlauthor{Ji Zhang}{damo}
\icmlauthor{Songfang Huang}{damo}
\icmlauthor{Fei Huang}{damo}
\icmlauthor{Jingren Zhou}{damo}

\end{icmlauthorlist}

\icmlaffiliation{damo}{DAMO Academy, Alibaba Group}
\icmlcorrespondingauthor{Ming Yan}{ym119608@alibaba-inc.com}

% You may provide any keywords that you
% find helpful for describing your paper; these are used to populate
% the "keywords" metadata in the PDF but will not be shown in the document
\icmlkeywords{Machine Learning, ICML}

\vskip 0.3in
]

% this must go after the closing bracket ] following \twocolumn[ ...

% This command actually creates the footnote in the first column
% listing the affiliations and the copyright notice.
% The command takes one argument, which is text to display at the start of the footnote.
% The \icmlEqualContribution command is standard text for equal contribution.
% Remove it (just {}) if you do not need this facility.

%\printAffiliationsAndNotice{}  % leave blank if no need to mention equal contribution
\printAffiliationsAndNotice{\icmlEqualContribution} % otherwise use the standard text.

\begin{abstract}
Recent years have witnessed a big convergence of language, vision, and multi-modal pretraining. In this work, we present \modelname, a new unified paradigm with modularized design for multi-modal pretraining, which can benefit from modality collaboration while addressing the problem of modality entanglement. In contrast to predominant paradigms of solely relying on sequence-to-sequence generation or encoder-based instance discrimination, \modelname introduces a multi-module composition network by sharing common universal modules for modality collaboration and disentangling different modality modules to deal with modality entanglement. It is flexible to select different modules for different understanding and generation tasks across all modalities including text, image, and video. Empirical study shows that \modelname achieves state-of-the-art or competitive results on a broad range of over 30 downstream tasks, spanning multi-modal tasks of image-text and video-text understanding and generation, and uni-modal tasks of text-only, image-only, and video-only understanding. Notably, \modelname shows new state-of-the-art results of 48.0 top-1 accuracy and 80.3 CIDEr on the challenging MSRVTT video QA and video caption tasks with a far smaller model size and data scale. It also demonstrates strong zero-shot transferability on vision-language and video-language tasks. Code and models will be released in \href{https://github.com/alibaba/AliceMind/tree/main/mPLUG}{https://github.com/alibaba/AliceMind}.

\end{abstract}

\vspace{-5ex}
\section{Introduction}
Large-scale pre-trained foundation models have been an emerging paradigm for a wide range of artificial intelligence (AI) fields, across language~\citep{devlin2018bert,Tom2020GPT3}, vision~\citep{dosovitskiy2020image,liu2021Swin} and multi-modality~\citep{radford2021learning,Yu2022CoCaCC,Wang2022BEITv3}. With the broad success of Transformer architecture~\citep{vaswani2017attention}, recent years have featured a trend toward the big convergence of language, vision and multimodal pre-training~\citep{Yu2022CoCaCC,Wang2022BEITv3,Alayrac2022FlamingoAV}. One line along this trend proposes to unify the tasks and modalities with a unified sequence-to-sequence generation framework such as T5~\citep{Colin2020T5}, OFA~\citep{Wang2022OFA} and Flamingo~\citep{Alayrac2022FlamingoAV}. On the other hand, BERT~\citep{devlin2018bert}, Florence~\citep{yuan2021florence} and BEIT-3~\citep{Wang2022BEITv3} models all the tasks as instance discrimination, and adopt the pure encoder-based architecture. 

%pre-training models have received tremendous success on a
%wide range of artificial intelligence (AI) fields, such as Natural Language Processing (NLP) (BERT \citep{}, T5 \citep{}, GPT3 \citep{}, ChatGPT\citep{}), Computer Vision (CV) (ViT \citep{}, BEIT \citep{}), Multimodality (VL-BERT \citep{}, CLIP \citep{}, DALL-E \citep{}). Recently, some foundation models (CoCa \citep{}, OFA \citep{}, Flamingo \citep{}) are further proposed to support the part of uni-modal and cross-modal tasks. However, these single models are difficult to support enough uni-modal and multi-modal tasks and achieve better performance at the same time. 

\begin{figure}
    \centering
    \includegraphics[width=1.1\linewidth]{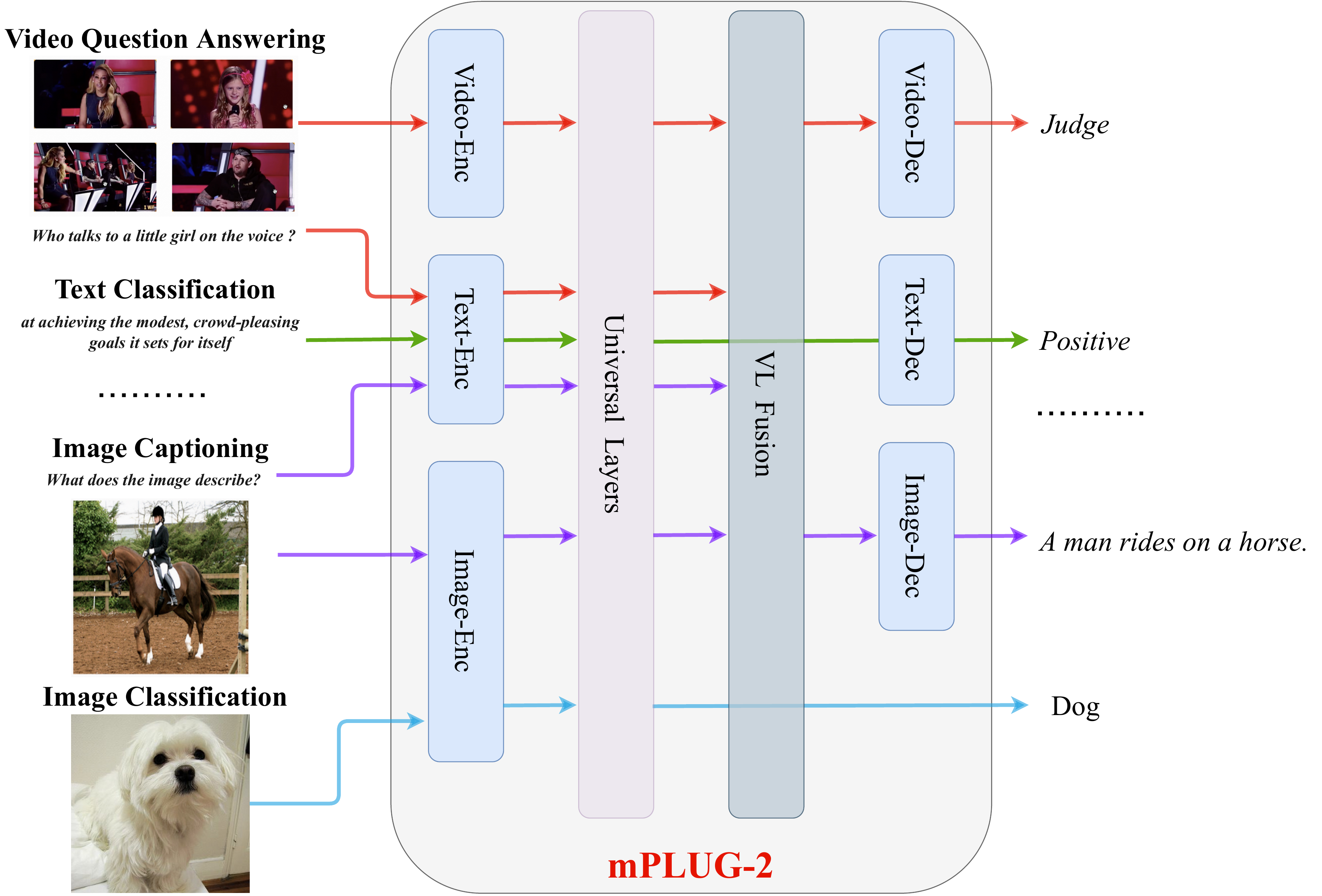}
\caption{A brief illustration of the new paradigm with modularized design for building multi-modal foundation model.}
\vspace{-4ex}
\label{fig:mplug2}
\end{figure}

\begin{table*}[t!]
\centering
\caption{\textbf{A system-level comparison between \modelname and existing foundation models in terms of various uni-modal and multi-modal downstream tasks.} "Cls." denotes the classification. "Det." and "Seg." are the short for "Detection" and "Segmentation" tasks respectively. "VG" stands for visual grounding task. Our \modelname is capable of supporting both uni-modal (i.e., CV and NLP) and multi-modal (i.e., Image-Text and Video-Text) downstream tasks simultaneously with the help of modularization.}
    \tablestyle{7pt}{1.1} 
    \def \w{15pt}
    \resizebox{0.95\linewidth}{!}{
    \begin{tabular}{l|cccc|ccc|cccc|ccc}
    \shline
              & \multicolumn{4}{c}{Computer Vision}   & \multicolumn{3}{c}{Natural Language Processing}  & \multicolumn{4}{c}{Image-Text} & \multicolumn{3}{c}{Video-Text} \\
    \cmidrule(lr){2-5} \cmidrule(lr){6-8} \cmidrule(lr){9-12} \cmidrule(lr){13-15}
    Method    & Image Cls. & Video Cls. & Det. & Seg. & Text Cls.                   & QA & Summarization & Retrieval  & QA  & Captioning & VG & Retrieval  & QA  & Captioning  \\
    \shline 
    BEiT-3      & $\checkmark$          &            & $\checkmark$    & $\checkmark$    &                             &    &               & $\checkmark$          & $\checkmark$   & $\checkmark$           &       &     &     &             \\
    EVA       & $\checkmark$          & $\checkmark$          & $\checkmark$    & $\checkmark$    &                             &    &               &            &     &             &      &      &     &             \\
    CLIP      & $\checkmark$          &            &      &      &                             &    &               & $\checkmark$          &     &          &    & $\checkmark$      &     &        \\
    ALBEF    &            &            &      &      &                             &    &               & $\checkmark$          & $\checkmark$   &            & $\checkmark$  &            &     &             \\
    BLIP      &            &            &      &      &                             &    &               & $\checkmark$          & $\checkmark$   & $\checkmark$       &    & $\checkmark$          & $\checkmark$   &             \\
    VATT      & $\checkmark$          & $\checkmark$          &      &      &                             &    &               &            &     &        &     & $\checkmark$          &     &             \\
    Florence  & $\checkmark$          & $\checkmark$          & $\checkmark$    &      &                             &    &               & $\checkmark$          & $\checkmark$   &            &  & $\checkmark$          &     &             \\
    CoCa      & $\checkmark$          & $\checkmark$          &      &      &                             &    &               & $\checkmark$          & $\checkmark$   & $\checkmark$           &  & $\checkmark$          &     &             \\
    VideoCoCa &            & $\checkmark$          &      &      &                             &    &               &            &     &             &    &        &     &             \\
    Flamingo  &            & $\checkmark$          &      &      &                             &    &               & $\checkmark$          & $\checkmark$   & $\checkmark$    &        &            & $\checkmark$   & $\checkmark$           \\
    GIT2      & $\checkmark$          &            &      &      & $\checkmark$                           &    &               &            & $\checkmark$   & $\checkmark$           &  &           & $\checkmark$   & $\checkmark$           \\
    FLAVA     & $\checkmark$          &            &      &      & $\checkmark$                           & $\checkmark$  & $\checkmark$             &            & $\checkmark$   &             &            &     &    &          \\
    OFA       & $\checkmark$          &            &      &      & $\checkmark$                           & $\checkmark$  & $\checkmark$             & $\checkmark$          & $\checkmark$   & $\checkmark$   & $\checkmark$        &            &     &             \\
    OmniVL    & $\checkmark$          & $\checkmark$          &      &      &                             &    &               & $\checkmark$          & $\checkmark$   & $\checkmark$ &           & $\checkmark$          & $\checkmark$   & $\checkmark$           \\
    \hline 
    mPLUG 2.0 & $\checkmark$          & $\checkmark$          & $\checkmark$    &  $\checkmark$   & $\checkmark$                           & $\checkmark$  &  $\checkmark$             & $\checkmark$          & $\checkmark$   & $\checkmark$    & $\checkmark$       & $\checkmark$          & $\checkmark$   & $\checkmark$        \\ 
    \shline
    \end{tabular}
    }
    \label{table:support-tasks}
    \vspace{-2ex}
\end{table*}

The predominant foundation models propose to share the same single network for multi-modality~\citep{Alayrac2022FlamingoAV} to leverage the information from modality collaboration. However, the strategy will suffer from the issue of modality entanglement due to the large variance of different modality tasks.
%However, due to the large variance of different modality tasks, the predominant foundation models that 
%benefit from modality collaboration by 
%share the same single network for multi-modality~\citep{Alayrac2022FlamingoAV}
%, while 
%will suffer from the issue of modality entanglement. 
The challenge is that multiple modalities may interfere with each other~\citep{huang2022modality}, especially when there are many modalities and tasks. It is difficult for a single-module foundation model to balance the gain of modality collaboration and the influence of modality entanglement on a large number of downstream tasks across multiple modalities. 

%there exists the problem of modality entanglement, where 

%We speculate that 1) there are the modality-pulling and modality-gap problems between different uni-modal and multi-modal datasets \citep{}; 2) various downstream tasks are difficult to depend on the single module such as retrieval, question answering, and captioning while achieving better performance simultaneously.

To alleviate the challenge, in this work, we introduce a new unified paradigm of multi-modal foundation models, as shown in Figure~\ref{fig:mplug2}. It features a module-based network design considering both the modality collaboration and modality entanglement, where \modelname designs certain shared functional modules to encourage the modality collaboration, while reserving modality-specific modules to tackle the problem of modality entanglement. Different modules are then jointly trained effectively on both the uni-modal and multi-modal datasets according to the task's module design. As a result, different modules can be flexibly selected and combined for the large number of uni-modal and cross-modal understanding and generation tasks accordingly. The details of the supported downstream tasks are given in Table~\ref{table:support-tasks}. To the best of our knowledge, the proposed method tackles the largest number of different kinds of downstream tasks across text, image and video.

\begin{figure*}
    \centering
    \includegraphics[width=0.95\linewidth]{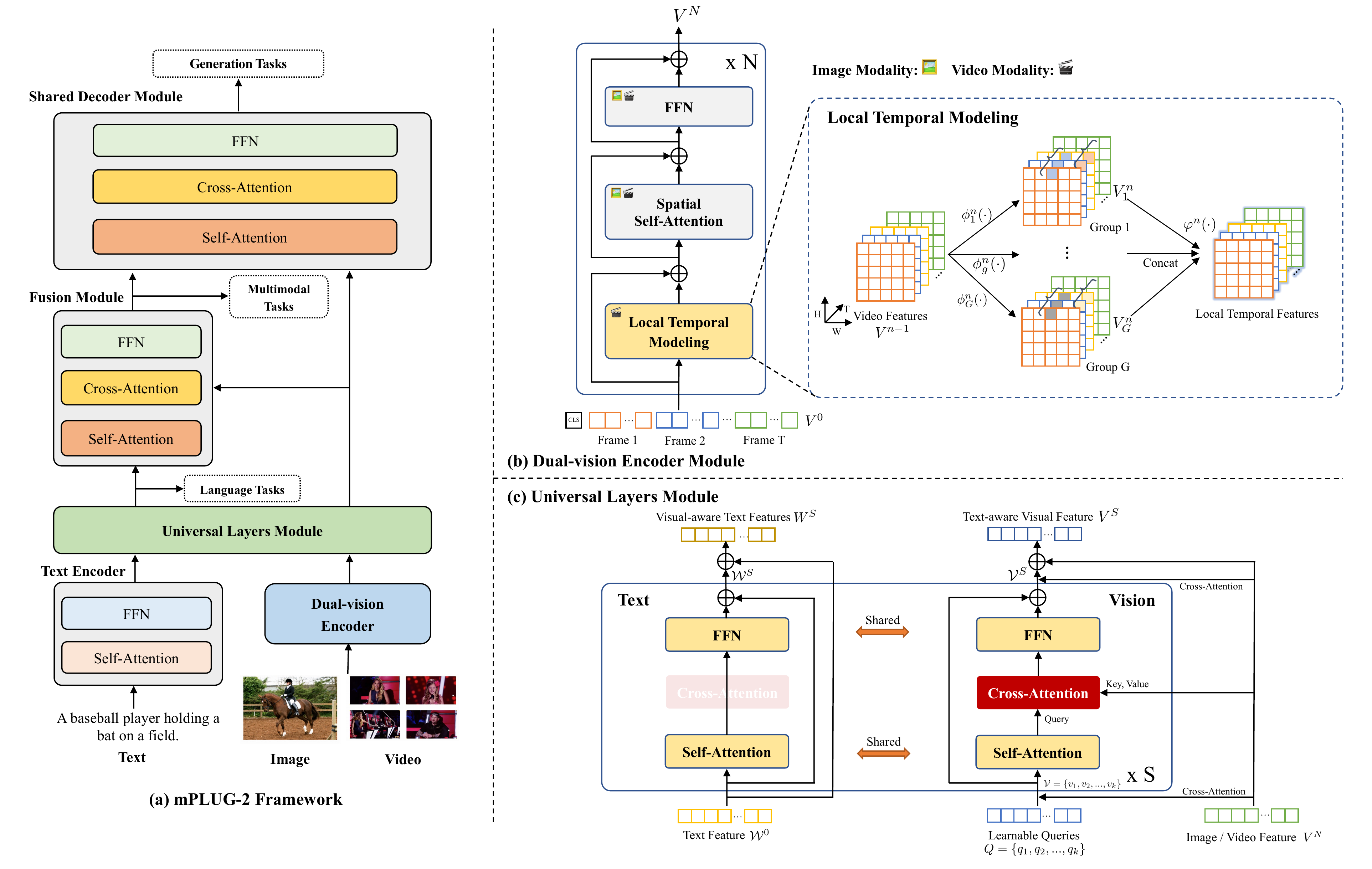}
\vspace{-2ex}
\caption{The overall framework and module details of \modelname.}
\vspace{-3ex}
\label{fig:model}
\end{figure*}

Specifically, we design a unified dual-vision encoder module by disentangling spatial and temporal representations, where video inputs share the standard Transformer module with image inputs for modeling spatial information and an extra local temporal modeling module is used for temporal relation modeling on video-related tasks. Then a novel universal layers module is introduced to serve as a pivot across different modalities, where vision and language modalities are projected to the common language-guided semantic space by sharing self-attention modules. Besides, an extra cross-attention module is used to fuse the universal vision representation with the original fine-grained vision representation. The detailed module design is shown in Figure~\ref{fig:model}. Finally, different modules of \modelname are jointly pre-trained with task and modality instructions~\citep{Wang2022OFA} on both uni-modal and cross-modal tasks. During inference, \modelname can select different modules for various uni-modal and cross-modal tasks with the modularized Transformer architecture. The selected modules for different tasks can be found in Table~\ref{table:module} in Appendix.

We evaluate the new unified paradigm of \modelname on over 30 challenging uni-modal and cross-modal understanding and generation benchmarks and it achieves state-of-the-art or competitive results with a similar model size and data scale. Equipping with the module-based network design, \modelname can be also easily extended to additional tasks by selecting and adding modules. Notably, \modelname shows new state-of-the-art results of 48.0 top-1 accuracy and 80.3 CIDEr on the challenging MSRVTT video QA and video caption tasks, respectively. \modelname also demonstrates strong zero-shot transferability on vision-language and video-language tasks. 

\vspace{-2ex}
\section{Related Work}
%\subsection{Foundation Models}
\paragraph{Vision-only Foundation Models}
ConvNets \citep{szegedy2017inception, He2015DeepRL} have long been the main stream visual architecture before the emergence of vision transformer (a.k.a. ViT) \citep{dosovitskiy2020image}. Due to the superior capacity of Transformer network, ViT stands out in various downstream tasks \citep{Nicolas2020DETR, Xu2022GroupViTSS}. Apart from scaling up the naive ViT architecture with large-scale dataset such as JFT-3B \citep{Zhai2021ScalingVT}, SwinV2-G \citep{Liu2021SwinTV} extends the original ViT with hierarchical architectures. 
% BEiT-3 \citep{Wang2022BEITv3} proposes a multi-way transformer \citep{vaswani2017attention} pre-trained with a unified mask data modeling task which enables the model to learn from images, texts, and multi-modal data simultaneously. 
In addition, EVA \citep{Fang2022EVA} distills the multi-modal knowledge to scale up ViT by leveraging unlabeled images with the large-scale pre-trained image-text model (e.g. CLIP \citep{radford2021learning}). Recently, InternImage \citep{Wang2022InternImageEL} revitalizes the convolutional neural networks with deformable convolution and achieves the state-of-the-art performance on various vision downstream tasks. Besides, InternVideo \citep{Wang2022InternVideoGV} extends to video tasks by assembling two large video models with both generative and discriminative self-supervised video learning.

\vspace{-2ex}
\paragraph{Language-only Foundation Models}
Inspired by the successful practice of the BERT~\citep{devlin2018bert} in natural language understanding, a massive large-scale language foundation models are proposed for natural language processing.
BART~\citep{Mike2020BART} is a denoising autoencoder like BERT but with encoder-decoder architecture which shows effectiveness for both text generation and comprehension tasks.
% BERT-series methods 的cite:
% cite: BERT, 
% cite: RoBERTa 
% BERT-series methods~\citep{} proved that pre-training with the masked language modeling objective helps the models learn syntactical and semantic knowledge.
Apart from BERT-series methods~\citep{devlin2018bert, Mike2020BART, liu2019roberta}, there are numerous other effective architectures and pre-training objectives. 
%XLNet~\citep{Zhilin2019XLNet} is based on Transformer-XL~\citep{Zihang2019Transformer-XL} and introduces the permutated language modeling objective to combine the advantages of autoregressive and autoencoding methods.
T5~\citep{Colin2020T5} introduce a unified framework that covers all text-based language tasks into a text-to-text format.
GPT-3~\citep{Tom2020GPT3} is an auto-regressive language foundation model which includes 175 billion parameters, and shows strong performance on many NLP tasks under the few-shot and zero-shot settings.
% ChatGPT: Optimizing Language Models for Dialogue
%Recently, ChatGPT% ~\citep{chatgpt}
%attracts wide attention by its amazing dialogue performance, which is fine-tuned from a model in the GPT-3.5 series and then trained this model using reinforcement learning from human feedback.

% XLNet
% BART
% T5
% GPT-2 GPT-3
\vspace{-2ex}
\paragraph{Vision-Language Foundation Models}
Benefiting from a large number of image/video-text pairs in the Internet, the emergence of vision-language foundation models can subsume vision-language pre-training. The success of CLIP \citep{radford2021learning} and ALIGN \citep{Jia2021ScalingUV} indicates that the model pre-trained with simple contrastive objectives on noisy image-text pairs can generate powerful vision-language representation. Moreover, ALBEF \citep{li2021align}, BLIP \citep{li2022blip} and mPLUG \citep{Li2022mPLUGEA} extend the task with multi-modal text completion and text generation for auxiliary learning. On the other hand, some foundation models are built through task unification. For instance, Florence \citep{yuan2021florence} unifies the contrastive objectives that can leverage both vision and vision-language data. BEiT-3 \citep{Wang2022BEITv3} ascribe the pre-training task to mask data modeling in terms of text, vision, and vision-language. SimVLM \citep{wang2021simvlm}, OFA \citep{Wang2022OFA}, and CoCa \citep{Yu2022CoCaCC} perform the generative pre-training for vision-language understanding and generation. 
%On the other hand, scaling up the model size can increase the models' capacity for powerful representation. For example, Flamingo \citep{Alayrac2022FlamingoAV} adopts a large frozen language model (4.8B in size) to keep the inherent few-shot learning ability, while GIT and GIT2 \citep{Wang2022GIT2} leverage a large contrastively pre-trained image encoder (12B in size) for powerful image representation. 
% Different from above foundation models, our proposed \modelname introduces modularized transformer framework with novel universal layers to project different modalities into common language-shared semantic space, which enables the model to perform well on more uni-modal and cross-modal tasks by utilizing different modules.
Different from predominant foundation models, \modelname introduces a new modularized transformer framework, which can leverage different compositions of modules for both uni-modal and cross-modal tasks by both sharing common universal modules and disentangling modality-specific ones to address the problem of modality entanglement.

%\subsection{Modularized Model}
% \begin{table}[t!]
% \centering
%     \tablestyle{7pt}{1.1} 
%     \def \w{15pt}
%     \resizebox{1.0\linewidth}{!}{
%     \begin{tabular}{l|ccc|cccccccc}
%         \shline
%         Tasks & Input & Modules \\
%         \shline 
%         Video-Text Retrieval & Video and Text & Video Enc+Text Enc+Universal Layers+Fusion Layers\\
%         Video-Text Question Answering & Video and Text & Video Enc+Text Enc
%         +Universal Layers+Fusion Layers+Video Dec \\
%         Video-Text Captioning & Video and Text  & Video Enc+Text Enc+Universal Layers+Fusion Layers+Video Dec \\
%         Image-Text Retrieval & Image and Text & Image Enc+Text Enc+Universal Layers+Fusion Layers\\
%         Image-Text Question Answering & Image and Text & Image Enc+Text Enc+Universal Layers+Fusion Layers+Image Dec \\
%         Image-Text Captioning & Image and Text & Image Enc+Text Enc+Universal Layers+Fusion Layers+Image Dec \\
%         Video Classification & Video & Video Enc+Universal Layers \\    
%         Image Classification & Image & Image Enc+Universal Layers \\
%         Image Detection & Image & Image Enc+Universal Layers \\
%         Image Segmentation & Image & Image Enc+Universal Layers \\
%         Text Classification & Text & Text Enc+Universal Layers \\
%         Text Question Answering & Text & Text Enc+Universal Layers \\
%         Text Summarization & Text & Text Enc+Universal Layers + Text Dec \\
%         \shline
%     \end{tabular}
%     }
%     \caption{\textbf{The modules of Tasks.
%     }}
%     \label{table:module}
% \end{table}
\begin{table*}
\centering
    \caption{\textbf{The modules for downstream tasks.
    }}
    \tablestyle{7pt}{1.1} 
    \def \w{15pt}
    \resizebox{1.0\linewidth}{!}{
    \begin{tabular}{l|ccc|cccccccc}
        \shline
         & \multicolumn{3}{c}{Input} & \multicolumn{8}{c}{Modules} \\
        \cmidrule(lr){2-4} \cmidrule(lr){5-12} 
        Tasks & Text & Image & Video & Text Enc & Image Enc & Video Enc & Universal Layers & Fusion Layers & Text Dec & Image Dec & Video Dec \\
        \shline 
        Video-Text Retrieval & \checkmark & & \checkmark  & \checkmark & & \checkmark& \checkmark & \checkmark &  & &\\
        Video-Text Question Answering & \checkmark & & \checkmark  & \checkmark & & \checkmark& \checkmark & \checkmark &  & &\checkmark \\
        Video-Text Captioning & \checkmark & & \checkmark  & \checkmark & & \checkmark& \checkmark & \checkmark &  & &\checkmark \\
        Image-Text Retrieval & \checkmark & \checkmark &   & \checkmark & \checkmark& & \checkmark & \checkmark &  & &\\
        Image-Text Question Answering & \checkmark & \checkmark &   & \checkmark & \checkmark& & \checkmark & \checkmark &  & \checkmark& \\
        Image-Text Captioning & \checkmark & \checkmark&  & \checkmark & \checkmark & & \checkmark & \checkmark &  &\checkmark & \\
        Visual Grounding & \checkmark & \checkmark &   & \checkmark & \checkmark& & \checkmark & \checkmark &  & \checkmark & \\
        Video Classification &  & & \checkmark &  & & \checkmark& \checkmark &  &  & & \\
        Image Classification &  & \checkmark &  &  & \checkmark & & \checkmark &  &  & &  \\
        Image Detection &  & \checkmark &  &  & \checkmark & & \checkmark &  &  & &  \\
        Image Segmentation &  & \checkmark &  &  & \checkmark & & \checkmark &  &  & &  \\
        Text Classification & \checkmark   & &  & \checkmark &  & & \checkmark &  &  & &  \\ 
        Text Question Answering & \checkmark   & &  & \checkmark &  & & \checkmark &  &  & &  \\ 
        Text Summarization & \checkmark  & &  & \checkmark &  & & \checkmark &  & \checkmark & &  \\ 
        \shline
    \end{tabular}
    }
    \label{table:module}
\end{table*}

\vspace{-3ex}
\section{Method}
\vspace{-1ex}
\subsection{Overall Framework}
As shown in Figure \ref{fig:model}, \modelname consists of a dual-vision encoder module for image and video, a text encoder module, a universal layers module that serves as a multi-modal pivot shared by all tasks, a multi-modal fusion module and a shared decoder module for uni-modal and cross-modal generation. We first use two uni-modal encoders which encode image/video and text separately to represent the inherent information of the individual modality. For image/video, we adopt the dual-vision encoder to encode visual features with spatial modeling and local temporal modeling. Then, the visual and linguistic representations are fed into the universal module separately, which consists of multiple universal layers. Each universal layer projects different modalities to shared semantic space for cross-modal alignment while preserving the original representation of different modalities. The output of universal layers is applied to conduct uni-modal discrimination tasks. For cross-modal tasks, an additional fusion module will be applied to produce cross-modal representations. Finally, the uni-modal and cross-modal representations can be incorporated as input to a shared Transformer decoder for various generation tasks, which facilitates multi-task pre-training and transfer learning. The modules for different downstream tasks are summarized in Table \ref{table:module}.

\vspace{-2ex}
\paragraph{Dual-vision Encoder Module}
To capture the visual information of various vision modalities, we propose dual-vision encoder to model image and video simultaneously. Specially, we split the image and video frames into a sequence of $L$ non-overlapping visual tokens. Every sequence of visual tokens with learnable spatial position embeddings and an extra [CLS] token constitute an input visual sequence.
%with learnable spatial position embeddings and an extra [CLS] token as the input visual sequence V^{0}V^{0}. 
% Then, the visual sequence is fed into a stack of NN transformer blocks for modeling the patch relations, which is computed as:
% \begin{align}
%     V_{SA}^{n} = LN(SA(V^{n-1}) + V^{n-1}), \\
%     V^{n} = LN(FFN(V_{SA}^{n}) + V_{SA}^{n}),
% \end{align}
% where n∈{1,⋯,N}n \in \{1, \cdots, N\}. The transformer blocks contains self-attention (SA) layer, the feed-forward network (FFN) and the layer normalization (LN).
However, modeling the completed visual sequences 
% brings large computation due to the lengthy sequence and 
leads to difficulty in spatio-temporal learning without large-scale video pre-training \citep{li2022lavender, wang2022allinone, Wang2022OmniVLOF}. To alleviate this problem, we decouple the visual representation into the spatial and temporal representation separately by introducing temporal locality.
As illustrated in Figure \ref{fig:model}(b), we leverage the self-attention (SA) layer and feed-forward layer (FFN) in the Transformer block for spatial modeling, and propose a novel local temporal modeling module (LT) to model the temporal dependency among the spatial representation as:
\begin{align}
    V_{LT}^{n} = LN(LT(V^{n-1}) + V^{n-1}), \\
    V_{SA}^{n} = LN(SA(V_{LT}^{n-1}) + V_{LT}^{n-1}), \\
    V^{n} = LN(FFN(V_{SA}^{n}) + V_{SA}^{n}), 
\end{align}
where LN is short for layer normalization. The local temporal modeling module captures the correlation among patches with the same spatial locations through multi-group fusion formulated as:
\begin{align}
    & V_g^{n} = ReLU(A_g^{n} \phi_g^{n} (V^{n-1})) \in \mathbb{R}^{T\times \frac{C}{G}} \\
    & LT(V^{n-1}) = \varphi^{n} (Concat[V_1^{n}; \cdots; V_G^{n}]),
\end{align}
where $\phi_g^n(\cdot)$ and $\varphi^{n}(\cdot)$ are linear transformation functions. $A_g^{n}$ is the learnable temporal relation parameter, which is instantiated as a convolution kernel. $T$ and $C$ are number of frames and size of hidden state. $G$ indicates the number of groups, and $Concat$ denotes concatenation function. By using multi-group fusion, the model is able to learn rich temporal information from distinctive representation subspaces at different temporal locations. As a result, except the local temporal module, the dual-vision encoder module enables weight sharing for images and videos, which effectively and efficiently learns the spatial and temporal representation.
% while significantly reducing the computation cost

\paragraph{Text Encoder Module}
For the text encoder module, we use BERT \citep{devlin2018bert} as the text encoder, which transforms the input text and an extra [CLS] token into a sequence of text embeddings. The embedding of [CLS] token is used to summarize the input text.

\paragraph{Universal Layers Module}
To benefit from modality collaboration, we propose the universal layers to model the vision and language modalities in the shared semantic space while preserving the original representation of the different modalities.

% Reason to project into language-shared space
% Therefore, the text modality, the universal layer serves as the plain transformer layers for the text token embeddings from text encoder module. 
Before the universal module, we take a variable number of image or video features $V^N$ from the dual-vision encoders as input to produce a fixed number $k$ of visual tokens $\mathcal{V}=\{\mathcal{}{v}_{1}, \mathcal{}{v}_{2}, ..., \mathcal{}{v}_{k}\} $ to reduce the computational complexity of universal layers.
% Similar to Flamingo \citep{Alayrac2022FlamingoAV} and DETR \citep{Nicolas2020DETR}, we learn a predefined number of latent input queries Q=\{q_1, q_2, ..., q_k\}Q=\{q_1, q_2, ..., q_k\}, which are fed to a cross-attention layer by the visual features. The universal layers modules consists of ss universal layers.
In the $i_{th}$ universal layer, the visual tokens $\mathcal{V}^{i-1}$ and the text representation $\mathcal{W}^{i-1}$ are fed to the shared self-attention layers to align semantics, and then the visual tokens are injected into the original visual feature space by the cross-attention layer to keep the original representation.
\begin{equation}
\mathcal{V}^{i}_{SA}=LN(SA(\mathcal{V}^{i-1})+\mathcal{V}^{i-1})
\end{equation}
\begin{equation}
\mathcal{W}^{i}_{SA}=LN(SA(\mathcal{W}^{i-1})+\mathcal{W}^{i-1})
\end{equation}
\begin{equation}
\mathcal{V}^{i}_{CA}=LN(CA(\mathcal{V}^{i}_{SA},V^{n})+\mathcal{V}^{i}_{SA})
\end{equation}
\begin{equation}
\mathcal{V}^{i}=LN(FFN(\mathcal{V}^{i}_{CA})+\mathcal{V}^{i}_{CA})
\end{equation}
\begin{equation}
\mathcal{W}^{i}=LN(FFN(\mathcal{W}^{i}_{SA})+\mathcal{W}^{i}_{SA})
\end{equation}
Then $[\mathcal{V}^{i}; \mathcal{W}^{i}]$ is fed into the next universal layer repeatedly to get the final common image and text representation. Finally, the output of the universal layers $[\mathcal{V}^{S}; \mathcal{W}^{S}]$ are combined with the original representations $[V^{N}; W^{M}]$ by the cross-attention layer for the text-aware visual and visual-aware text representation, where $S, N, M$ are the layers of universal module, dual-vision encoder and text encoder respectively.

\begin{table*}[t!]
\centering
\caption{\textbf{Performance comparison on text-to-video retrieval.} All results are reported on R@1/R@5/R@10. 
    }
    \tablestyle{7pt}{1.1} 
    \def \w{15pt}
    \resizebox{0.8\linewidth}{!}{
    \begin{tabular}{ll|ccc|ccc|ccc}
        \shline
        ~ & ~ & \multicolumn{3}{c}{MSRVTT}& \multicolumn{3}{c}{DiDeMo} & \multicolumn{3}{c}{LSMDC} \\
        \cmidrule(lr){3-5} \cmidrule(lr){6-8} \cmidrule(lr){9-11}
        Method & \#PT Data & R@1 & R@5 & R@10 & R@1 & R@5 & R@10 & R@1 & R@5 & R@10 \\
        \shline 
        Frozen~\citep{bain2021frozen} & 5M  & 31.0 & 59.5 & 70.5 & 31.0 & 59.8 & 72.4 & 15.0 & 30.8 & 39.8 \\
        BridgeFormer~\citep{ge2022bridgeformer} & 5M  & 37.6 & 64.8 & 75.1 & 37.0 & 62.2 & 73.9  & 17.9 & 35.4 & 44.5 \\
        Singularity~\citep{lei2022singularity} & 5M & 36.8 & 65.9 & 75.5 & 47.4 & 75.2 & 84.0 & - & - & - \\
        LAVENDER~\citep{li2022lavender} & 30M & 37.8 & 63.8 & 75.0 & 47.4 & 74.7 & 82.4 & 22.2 & 43.8 & 53.5 \\
        All-in-one~\citep{wang2022allinone} & 283M  & 37.9 & 68.1 & 77.1 & 32.7 & 61.4 & 73.5 & - & - & -\\
        % Clip4Clip~\citep{} & 400M & 42.1 & 71.9 & 81.4 & 43.4 & 70.2 & 80.6 & 21.6 & 41.8 & 49.8 \\
        % X-CLIP~\citep{} & 400M & 46.1 & 73.0 & 83.1 & 45.2 & 74.0 & -  & 23.3 & 43.0 & - \\
        OmniVL~\citep{Wang2022OmniVLOF} & 18M & 47.8 & 74.2 & 83.8 & 52.4 & 79.5 & 85.4 & - & - & - \\
        HiTeA~\citep{Ye2022HiTeAHT} & 17M & 46.8 & 71.2 & 81.9 & \textbf{56.5} & \textbf{81.7} & \textbf{89.7} & 28.7 & 50.3 & 59.0 \\
        % CLIP-ViP~\citep{} & 100M & 54.2 & 77.2 & 84.8 & 50.5 & 78.4 & 87.1 & 29.4 & 50.6 & 59.0 \\ 
        \hline 
        \modelnamebase & 17M & 48.3 & 75.0 & 83.2 & 52.3 & 80.8 & 87.5 & 25.5 & 45.8 & 55.8 \\
        \modelname & 17M & \textbf{53.1} & \textbf{77.6} & \textbf{84.7} & 56.4 & 79.1 & 85.2 & \textbf{34.4} & \textbf{55.2} & \textbf{65.1} \\
        \shline
    \end{tabular}
    }
    
    \label{table:video-retrieval}
    \vspace{-2ex}
\end{table*}

% \begin{table}[t!]
% \centering
%     \tablestyle{7pt}{1.1} 
%     \def \w{15pt}
%     \resizebox{1.0\linewidth}{!}{
%     \begin{tabular}{ll|ccc}
%         \shline
%         Method & Params & MSRVTT-QA & MSVD-QA & TGIF-Frame \\
%         \shline 
%         JustAsk~\citep{} & xxx & 51.5 & 46.3 & - \\
%         LAVENDER~\citep{} & xx & 45.0 & 56.6 & 73.5 \\
%         All-in-one~\citep{} & xxx & 46.8 & 48.3 & 66.3 \\
%         MERLOT~\citep{} & xxx & 43.1 & - & 69.5 \\
%         OmniVL~\citep{} & xxx & 44.1 & 51.0 & - \\
%         HiTeA~\citep{} & xxx & 45.9 & 55.3 & 73.2 \\
%         GIT~\citep{} & 0.7B & 43.2 & 56.8 & 72.8 \\
%         GIT2~\citep{} & 5.1B & 45.6 & 58.2 & 74.9 \\
%         FrozenBiLM~\citep{} & xxx & 47.0 & 54.8 & 68.6 \\
%         VideoCoca~\citep{} & 2.1B & 46.0 & 56.9 & - \\
%         InternVideo~\citep{} & 1.3B & 47.1 & 55.5 & 72.2 \\
%         Flamingo~\citep{} & 80B & 47.4 & - & - \\
%         \hline 
%         mPLUG 2.0 & xxx & xxx & xxx & xxx \\
%         mPLUG 2.0 & xxx & xxx & xxx & xxx \\
%         \shline
%     \end{tabular}
%     }
%     \caption{\textbf{Performance comparison on video question answering.} Accuracy is reported for evaluation.
%     }
%     \label{table:video-qa}
% \end{table}

\begin{table}[t!]
\caption{\textbf{Performance comparison on video question answering.} Accuracy is reported for evaluation. \modelname creates a new state-of-the-art video question answering results on MSRVTT-QA and TGIF-FrameQA with open-vocabulary generation.
    }
\centering
    \tablestyle{7pt}{1.1} 
    \def \w{15pt}
    \resizebox{1.0\linewidth}{!}{
    \begin{tabular}{ll|ccc}
        \shline
        Method & \#PT Data & MSRVTT-QA & MSVD-QA & TGIF-FrameQA \\
        \shline 
        JustAsk~\citep{yang2021justask} & 69M & 41.5 & 46.3 & - \\
        LAVENDER~\citep{li2022lavender} & 30M & 45.0 & 56.6 & 73.5 \\
        All-in-one~\citep{wang2022allinone} & 283M & 46.8 & 48.3 & 66.3 \\
        MERLOT~\citep{zellers2021merlot} & 180M & 43.1 & - & 69.5 \\
        OmniVL~\citep{Wang2022OmniVLOF} & 18M & 44.1 & 51.0 & - \\
        HiTeA~\citep{Ye2022HiTeAHT} & 17M & 45.9 & 55.3 & 73.2 \\
        GIT~\citep{Wang2022GIT2} & 800M & 43.2 & 56.8 & 72.8 \\
        GIT2~\citep{Wang2022GIT2} & 12.9B & 45.6 & \textbf{58.2} & 74.9 \\
        FrozenBiLM~\citep{yang2022frozenbilm} & 10M & 47.0 & 54.8 & 68.6 \\
        VideoCoCa~\citep{Yan2022VideoCoCa} & 3B & 46.0 & 56.9 & - \\
        InternVideo~\citep{Wang2022InternVideoGV} & 12M & 47.1 & 55.5 & 72.2 \\
        Flamingo~\citep{Alayrac2022FlamingoAV} & 2.3B & 47.4 & - & - \\
        \hline 
        \modelnamebase & 17M & 46.3 & 55.3 & 72.6 \\
        \modelname & 17M & \textbf{48.0} & 58.1 & \textbf{75.4} \\
        \shline
    \end{tabular}
    }

    \label{table:video-qa}
    \vspace{-2ex}
\end{table}

\paragraph{Fusion Module}
To effectively capture the cross-modal interaction between vision and language modalities, we use the fusion module as in ALBEF~\citep{li2021align}, which is composed of a stack of Transformer blocks with cross-attention layers. Specifically, the fusion module takes the text embeddings from the universal layers module as the input. Then, the text-aware vision embedding cross-attends to the visual-aware text embeddings in language-shared common space. By cascading the Transformer blocks with cross-attention layers, fusion module is able to yield multi-modal vision-language representations.

\vspace{-2ex}
\paragraph{Shared Decoder Module}
To empower the model with the capability of generation, a shared decoder module is introduced to enable the model to generate text with both uni-modal and multi-modal information. In detail, the shared decoder module is a Transformer decoder with arbitrary inputs. For example, image captioning only requires the visual features, while the multi-modal features are used for visual question answering. By taking different types of input, our shared decoder module can adapt to a variety of tasks with text generation. The shared decoder module facilitates multi-task pre-training and transfer learning.

\vspace{-2ex}
\subsection{Unified Pre-training Objectives}
We jointly train the multiple modules of \modelname with the following three objectives.

\vspace{-2ex}
\paragraph{Language Loss}
For the text encoder module, we use Masked Language Modeling (MLM) as in BERT \citep{devlin2018bert} to learn the text representation. We randomly mask 15\% tokens in the text and the model is asked to predict these masked tokens with the context representations.
\vspace{-2ex}
\paragraph{Multi-modal Loss}
For the cross-modal module, we employ the Cross-modal Matching Losses (CML) as in ALBEF \citep{li2021align}, which consists of Vision-language Matching (VLM) and Vision-language Contrastive Learning (VLC). 
\vspace{-2ex}
\paragraph{Instruction-based Language Model Loss}
Following Flamingo \citep{Alayrac2022FlamingoAV} and OFA \citep{Wang2022OFA}, we adopt the Instruction-based Language Model Loss to unify various generation tasks. We use handcrafted instructions to discriminate tasks and modalities, which include Video/Image-Text Pairs, Video/Image Captioning, Video/Image Question Answering, Text Generation, etc. 

\vspace{-2ex}
\section{Experiment}
\subsection{Training Setup}
\paragraph{Pre-training Datasets}
Following previous works \citep{li2021align, Li2022mPLUGEA}, we pre-train our model with the same popular image-text datasets with 14M images including MS COCO \citep{lin2014microsoft}, Visual Genome \citep{krishna2017visual}, Conceptual Captions 3M \citep{sharma2018conceptual}, Conceptual Captions 12M \citep{changpinyo2021conceptual}, and SBU Captions \citep{Ordonez2011Im2TextDI}. For video-text datasets, we adopt the web-sourced video dataset WebVid-2M \citep{Bain2021FrozenIT} with 2.5M video-text pairs. The text datasets consists of WikiCorpus \citep{devlin2018bert} (about 20GB) and cleaned common crawl (about 350GB). The collection and cleaning method of the latter is generally the same as that used in c4 \citep{Colin2020T5}. The implementation details of pre-training can be found in the Appendix.

\vspace{-2ex}
\subsection{Main Results}

\begin{table*}[t!]
\centering
\caption{\textbf{Performance comparison on video captioning.} B@4: BLEU@4; M: METEOR; R: ROUGE-L; C: CIDEr.
    }
    \tablestyle{7pt}{1.1} 
    \def \w{15pt}
    \resizebox{0.7\linewidth}{!}{
    \begin{tabular}{ll|cccc|cccc}
        \shline
        ~ & ~ & \multicolumn{4}{c}{MSRVTT}& \multicolumn{4}{c}{MSVD} \\
        \cmidrule(lr){3-6} \cmidrule(lr){7-10}
        Method & \#PT Data & B@4 & M & R & C & B@4 & M & R & C  \\
        \shline
        UniVL \citep{luo2020univl} & 136M & 42.2 & 28.2 & 61.2 & 49.9 & - & - & - & - \\
        SwinBERT \citep{lin2022swinbert} & - & 41.9 & 29.9 & 62.1 & 53.8 & 58.2 & 41.3 & 77.5 & 120.6  \\
        CLIP4Caption \citep{tang2021clip4caption} & - & 46.1 & 30.7 & 63.7 & 57.7 & - & - & - & - \\
        MV-GPT \citep{seo2022mvgpt} & 69M & 48.9 & 38.7 & 64.0 & 60.0 & - & - & - & - \\
        LAVENDER \citep{li2022lavender} & 30M &  - & - & - & 60.1 & - & - & - & 150.7 \\
        HiTeA \citep{Ye2022HiTeAHT} & 17M & 49.2 & 30.7 & 65.0 & 65.1 & 71.0 & 45.3 & 81.4 & 146.9 \\
        VideoCoca \citep{Yan2022VideoCoCa} & 3B & 53.8 & - & 68.0 & 73.2 & - & - & - & - \\
        GIT \citep{Wang2022GIT2} & 0.8B & 53.8 & 32.9 & 67.7 & 73.9 & 79.5 &  51.1 &  87.3 &  180.2 \\
        GIT2 \citep{Wang2022GIT2} & 12.9B & 54.8 & 33.1 & 68.2 & 75.9 & \textbf{82.2} & \textbf{52.3} & \textbf{88.7} & \textbf{185.4} \\
        \hline
        \modelnamebase & 17M & 52.2 & 32.1 & 66.9 & 72.4 & 69.3 & 45.1 & 81.9 & 148.2 \\
        % mPLUG 2.0 & 17M & - & - & - & - & 75.0 & 48.4 & 85.3 & 165.8 \\
        \modelname & 17M & \textbf{57.8} & \textbf{34.9} & \textbf{70.1} & \textbf{80.3} & 75.0 & 48.4 & 85.3 & 165.8 \\
        % \demph{\textbf{HiTeA}*} & \demph{17M} & \demph{\textbf{73.1}} & \demph{\textbf{157.2}} \\
        % \modelname & 5M & \textbf{xxx} & \textbf{xxx} \\
        \shline
    \end{tabular}
    }
    \label{table:video-caption}
    \vspace{-2ex}
\end{table*}
\begin{table*}
\setlength\tabcolsep{4pt}
\centering
\caption{\textbf{Performance comparison on image-text retrieval.} All results are reported on R@1/R@5/R@10. }

\small

\resizebox{0.8\linewidth}{!}{
\begin{tabular}{ll|cccccc|cccccc}
\shline
% \multicolumn{1}{c|}{\multirow{2}{*}{Models}}      &
% %\multirow{2}{*}{\# Pretrain} &
% \multicolumn{1}{c|}{\# Pretrain} &
% \multicolumn{6}{c|}{MSCOCO (5K test set)} & \multicolumn{6}{c}{Flickr30K (1K test set)} \\
%       &  data & \multicolumn{3}{c}{TR} & \multicolumn{3}{c|}{IR} & \multicolumn{3}{c}{TR} & \multicolumn{3}{c}{IR}          \\
% \midrule

~ & ~ & \multicolumn{6}{c}{MSCOCO (5K test set)}& \multicolumn{6}{c}{Flickr30K (1K test set)} \\
        % \cmidrule(lr){3-8} \cmidrule(lr){9-14}
~ & ~ & \multicolumn{3}{c}{TR} & \multicolumn{3}{c}{IR} & \multicolumn{3}{c}{TR} & \multicolumn{3}{c}{IR} \\
        \cmidrule(lr){3-5} \cmidrule(lr){6-8} \cmidrule(lr){9-11} \cmidrule(lr){12-14}
Method & \#PT Data & R@1&R@5&R@10&R@1&R@5&R@10&R@1&R@5&R@10&R@1&R@5&R@10 \\
\shline
E2E-VLP~\citep{xu2021e2e}& 4M     &-& -&-&-&-&- & 86.2 &97.5 &98.92&73.6 & 92.4 &96.0 \\
UNITER~\citep{chen2020uniter} & 4M     & 65.7&88.6&93.8&52.9&79.9&88.0&87.3& 98.0&99.2&75.6&94.1&96.8  \\
OSCAR~\citep{li2020oscar} & 4M  & 70.0&91.1&95.5&54.0&80.8&88.5&-& -&-&-&-&-   \\
UNIMO ~\citep{li2020unimo} & 4M     &-& -&-&-&-&- & 89.4 & 98.9& 99.8 &78.0 &94.2& 97.1\\
VLMo~\citep{wang2021vlmo} & 4M & 78.2& 94.4& 97.4& 60.6& 84.4& 91.0& 95.3& 99.9& \textbf{100.0}& 84.5& 97.3& 98.6 \\
ALIGN~\citep{Jia2021ScalingUV} & 1.8B  & 77.0&93.5&96.9&59.9&83.3&89.8&95.3& 99.8&\textbf{100.0}&84.9&97.4&98.6   \\
ALBEF~\citep{li2021align} & 14M & 77.6&94.3&97.2&60.7&84.3&90.5&95.9& 99.8&\textbf{100.0}&85.6&97.5&98.9                 \\
Florence~\citep{yuan2021florence} & 0.9B & 81.8&95.2&-&63.2&85.7&-&97.2& 99.9&-&87.9&\textbf{98.1}&-                 \\
% BLIP~\citep{li2022blip}& 14M & 80.6 &95.2&97.6&63.1&85.3&91.1&96.6& 99.8&100.0&87.2&97.5&98.8                 \\
BLIP~\citep{li2022blip}& 129M & 82.4 &95.4&97.9&65.1&86.3&91.8&\textbf{97.4}& 99.8&99.9&87.6&97.7&99.0                 \\
\hline
\modelnamebase & 17M & 81.2 & 95.6 & \textbf{98.1} & 65.3 & 86.9 & 92.4  & 96.9 & \textbf{100.0} & \textbf{100.0} & \textbf{88.2} & 97.8 & 99.0  \\
\modelname & 17M & \textbf{82.5} & \textbf{95.7} & 98.0 & \textbf{65.7} & \textbf{87.1} & \textbf{92.6}  & 97.2 & \textbf{100.0} & \textbf{100.0} & 88.1 & 97.6 & \textbf{99.1}  \\
\shline
\end{tabular} 
}
\label{table:retrieval}
\vspace{-2ex}
\end{table*}
We evaluate the new unified paradigm of mPLUG-2 on over 30
benchmarks including vision-language tasks (e.g. multi-modal retrieval, question answering and captioning) \citep{xu2016msrvtt, xu2017msrvttqa, chen2011msvd}, language-only tasks (e.g. text classification, question answering and summarization) \citep{Wang2018GLUEAM, Rush2015ANA}, and vision-only tasks (e.g. image classification and video action recognition) \citep{imagenet, Kay2017Kinetics}. Specially, the vision-language benchmarks can be categorized as image-text parts and video-text parts. Details of these datasets can be found in the Appendix.

\vspace{-2ex}
\subsubsection{Multi-modal Tasks}
% We first evaluate the video-language performance of \modelname on several video-text downstream tasks including text-to-video retrieval, video question answering, and video captioning. Then we examine that our method is capable of handling the image-text tasks simultaneously. In detail, we perform image-text retrieval, visual question answering and image captioning for evaluation image-text understanding and generation capabilities.

\vspace{-1ex}
\paragraph{Text-to-video Retrieval}
We compare \modelname with several state-of-the-art methods on MSRVTT \citep{xu2016msrvtt}, DiDeMo \citep{anne2017didemo} and LSMDC \citep{rohrbach2015lsmdc} datasets. The results are summarized in Table~\ref{table:video-retrieval}. We can observe that \modelname outperforms the previous SoTA methods on most of the datasets. In particular, our method yields 5.7\% lift in terms of R@1 on LSMDC datasets compared with HiTeA, which indicates that the proposed model can leverage the temporal information presented in fruitful movie clips through the proposed local temporal modeling module in the dual-vision encoder.

\paragraph{Video Question Answering}
Table \ref{table:video-qa} summarizes the video question answering results on MSRVTT-QA \citep{xu2017msrvttqa}, MSVD-QA \citep{xu2017msrvttqa}, and TGIF-FrameQA \citep{jang2017tgif}. It can be observed that \modelname outperforms all the existing foundation models on MSRVTT-QA and TGIF-FrameQA by a large margin, and it also attains the comparable result with big foundation models GIT2 \citep{Wang2022GIT2} on MSVD-QA even using significantly smaller amount of pre-trained data. In particular, \modelname achieves absolute improvement 0.6\% on MSRVTT and 0.5\% on TGIF-FrameQA. Furthermore, \modelnamebase achieves the comparable results compared to the large models (i.e., VideoCoCa and GIT2) with smaller model size.

\paragraph{Video Captioning} 
Table \ref{table:video-caption}\ref{table:video-caption} compares \modelname with existing methods on video captioning datasets MSRVTT and MSVD. As shown in the table, although pre-trained on less data, \modelname derives the significant improvement on MSRVTT dataset, and comparable performance on MSVD dataset. On MSRVTT Caption, our method surpasses SoTA method VideoCoCa \citep{Yan2022VideoCoCa} and GIT2 \citep{Wang2022GIT2} by 4.4\% on CIDEr and 3.0\% on BLEU@4. Moreover, we can notice \modelname outperforms HiTeA with the same amount of pre-training data, which shows that \modelname is able to generate stronger video-language representation.

\paragraph{Visual Grounding}
\begin{table}[t]
\small
\centering
\caption{\textbf{Evaluation results on visual grounding (ReferCOCO and ReferCOCOg).} We use the accuracy@0.5 (a prediction is right if the IoU between the grounding-truth box and the predicted bounding box is larger than 0.5) to measure model performance.}
\label{tab:visual_grounding_single}
\resizebox{\linewidth}{!}{
\begin{tabular}{@{}lccccccccc@{}}
\toprule
\multicolumn{1}{c}{\multirow{2}{*}{Model}} & \multicolumn{3}{c}{RefCOCO} & \multicolumn{2}{c}{RefCOCOg} \\
\multicolumn{1}{c}{}                       & val     & testA   & testB &  val-u        & test-u       \\ \midrule
UNITER~\citep{chen2020uniter}  & 81.41   & 87.04   & 74.17    & 74.86 & 75.77         \\
VILLA~\citep{Zhe2020VILLA}   & 82.39   & 87.48   & 74.84    & 76.18        & 76.71        \\
MDETR~\citep{kamath2021mdetr}   & 86.75   & 89.58   & 81.41   & 81.64      & 80.89  \\
UNICORN~\citep{Yang2021UNICORN}  & 88.29   & 90.42   & 83.06     & 83.44   & 83.93       \\
OFA$_{Large}$~\citep{Wang2022OFA}  & 90.05   & \textbf{92.93}   & 85.26    & 84.54        & \textbf{85.20}   \\ 
%mPLUG 1.0                                 & \textbf{92.40}   & \textbf{94.51}   & \textbf{88.42}   &  \textbf{86.02}   &  \textbf{90.17}   &  \textbf{78.17}   &    \textbf{85.88}   & \textbf{86.42}   \\ 
\midrule
\modelname                                 & \textbf{90.33}   & 92.80  & \textbf{86.05} & \textbf{84.70}   & 85.14   \\ \bottomrule
\end{tabular}
}

\end{table}
We compare \modelname with existing state-of-the-art methods on visual grounding datasets including RefCOCO~\citep{yu2016modeling}, RefCOCO+~\citep{yu2016modeling} and RefCOCOg~\citep{mao2016generation}. \cref{tab:visual_grounding_single} shows that \modelname achieves comparable performance to the state-of-the-art methods. Our method achieve 0.97\% absolute improvement compared with the second best method on RefCOCO ``testB'' split without using object detection data for pre-training. Queries in ``testB'' split can refer to various visual concepts but only people in ``testA''. The improvement demonstrates that the introduction of universal layers can help model the visual concepts in the image.

\paragraph{Image-Text Retrieval}
\vspace{-2ex}
We evaluate \modelname on image-text retrieval datasets MSCOCO and Flickr30k. As shown in \Cref{table:retrieval}, both \modelnamebase and \modelname achieves comparable or better performance than state-of-the-art methods. Florence~\citep{yuan2021florence} and BLIP~\citep{li2022blip} use 0.9B and 129M data for pre-train respectively. In contrast, our \modelname only requires 17M data. It demonstrate that \modelname is data-efficient.
\paragraph{Visual Question Answering}
\begin{table}[t]
% \caption{Comparison to existing methods on VQA on test sets.}
\caption{\textbf{Performance comparison on visual question answering.} Accuracy is reported for evaluation.}
\tablestyle{7pt}{1.05}
\resizebox{0.85\linewidth}{!}{
\begin{tabular}{ll|cc}
\shline
Method & \#PT Data & test-dev & test-std \\
\shline
UNITER~\citep{chen2020uniter} & 4M & 72.70 & 72.91 \\
UNIMO~\citep{li2020unimo} & 4M & 73.79 & 74.02 \\
E2E-VLP~\citep{xu2021e2e} & 4M & 73.25  & 73.67 \\
OSCAR~\citep{li2020oscar} & 4M & 73.16 & 73.44 \\
ALBEF~\citep{li2021align} & 4M & 74.54 & 74.70 \\
BLIP~\citep{li2022blip} & 14M & 77.54 & 77.62 \\
SimVLM~\citep{wang2021simvlm} & 1.8B  &  80.03 & 80.34 \\
Florence~\citep{yuan2021florence} & 0.9B & 80.16 & 80.36\\
OFA$_{Large}$~\citep{Wang2022OFA} & 18M & 80.30 & 80.50 \\
VLMo~\citep{wang2021vlmo} & - & 79.94 & 79.98 \\
GIT \citep{Wang2022GIT2} & 0.8B & 78.56 & 78.81 \\
\shline
\modelnamebase & 17M & 79.27 & 79.32 \\
\modelname & 17M & \textbf{81.11} & \textbf{81.13} \\
\shline
\end{tabular}
}
\label{tab:vqa}
\vspace{-2ex}
\end{table}

\begin{table}
\centering
% \caption{Evaluation Results on COCO Caption "Karpathy" test split and NoCaps validation set. B@4: BLEU@4, M: METEOR, C: CIDEr, S: SPICE.} 
\caption{\textbf{Performance comparison on image captioning.} B@4: BLEU@4; M: METEOR; R: ROUGE-L; C: CIDEr.
    }
\resizebox{0.99\linewidth}{!}{
\begin{tabular}{lc|cccc}
\shline
% \multicolumn{1}{c|}{\multirow{3}{*}{Models}}      &
% %\multirow{2}{*}{\# Pretrain} &
% \multicolumn{1}{c|}{\multirow{3}{*}{Data}} &
% \multicolumn{8}{c|}{COCO Caption} & \multicolumn{2}{c}{\multirow{1}{*}{NoCaps}}  \\
% \multicolumn{1}{c|}{\multirow{2}{*}{}}      &
% % \multirow{2}{*}{\# Pretrain} &
% \multicolumn{1}{c|}{} &
% \multicolumn{4}{c}{Cross-entropy Optimization} & \multicolumn{4}{c|}{CIDEr Optimization} & \multicolumn{2}{c}{}  \\
%       &  & B@4 & M & C & S & B@4 & M & C & S & C & S     \\

~ & ~ & \multicolumn{4}{c}{COCO Caption}  \\
        \cmidrule(lr){3-6}
Method & \#PT Data & B@4 & M & C & S  \\
\shline      
Encoder-Decoder & 12M & - & - & 110.9 & - \\
E2E-VLP \citep{xu2021e2e} & 4M & 36.2 & - &117.3& -  \\
VinVL \citep{zhang2021vinvl} & 5.65M & 38.5 & 30.4 & 130.8  \\
OSCAR \citep{li2020oscar} & 6.5M & - & - & - & -  \\
% SimVLM_{large}_{large} \citep{wang2021simvlm} & 1.8B & 40.3 & \textbf{33.4} & \textbf{142.6} & \textbf{24.7} & - & - \\
% SimVLMhuge_{huge} \cite{wang2021simvlm} & 1.8B & 40.6 & 33.7 & 143.3 & 25.4 & - & - & - & - & 112.2 & -  \\
LEMON$_{large}$ \citep{LEMON} & 200M & 40.6 & 30.4 & 135.7 & 23.5  \\
BLIP \citep{li2022blip} & 129M & 40.4 & - & 136.7 & -  \\
% OFA \cite{wang2022OFA} & 18M & - & - & - & - & 43.5 & 31.9 & 149.6 & \textbf{26.1} & - & - \\
% mPLUG \citep{} & 14M  & 43.1 & 31.4 & 141.0 & 24.2 & 114.8 & 14.8 \\
\hline
\modelname & 17M  & \textbf{41.6} & \textbf{30.9} & \textbf{137.7} & \textbf{23.7} \\

\shline

\end{tabular}
}
\label{table:image-caption}
\vspace{-2ex}
\end{table}
We report the performance of \modelname on visual question answering test sets. \modelname surpasses state-of-the-art method Florence \citep{yuan2021florence} 0.95\% on test-dev and 0.77\% on test-std. The scale of the pre-trained data used in our model is 89.11\% less than that in Florence. It shows that our \modelname can learn multi-modal represent efficiently and effectively.
\paragraph{Image Captioning}
% We compare mPLUG 2.0 with existing state-of-the-art methods on MSCOCO~\citep{lin2014microsoft} and NoCaps~\citep{nocaps}. 
% % following oscar
% Following~\citep{li2020oscar}, we train \modelname on the COCO Caption with cross-entropy loss and test on the same Karpathy split and NoCaps validation set.
%\our \modelname achieves new SoTA results on COCO Caption. 
% Moreover, our method achieves competitive results with big foundation models, such as LEMON~\citep{LEMON} and BLIP~\citep{li2022blip} which use more than nearly 10x amount of pre-training data.
% Specifically, our \modelname outperforms BLIP on COCO caption by an obvious 1.2 point margin on BLEU@4, and achieves comparable results on NoCaps. 

We compare \modelname with existing state-of-the-art methods on MSCOCO~\citep{lin2014microsoft}. 
% following oscar
Following~\citep{li2020oscar}, we train \modelname on the COCO Caption with cross-entropy loss and test on the same Karpathy split.
As shown in Table \ref{table:image-caption}, our \modelname achieves new SoTA results on COCO Caption. 
Moreover, our method achieves competitive results with big foundation models, such as LEMON~\citep{LEMON} and BLIP~\citep{li2022blip} which use more than nearly 10x amount of pre-training data.
Specifically, our \modelname outperforms BLIP on COCO caption by an obvious 1.2 point margin on BLEU@4, and 1 point on CIDEr. 

% \paragraph{Visual Grounding}
% \input{tables/visual_grounding_single.tex}
% We compare \modelname with existing state-of-the-art methods on visual grounding datasets including RefCOCO~\citep{yu2016modeling}, RefCOCO+~\citep{yu2016modeling} and RefCOCOg~\citep{mao2016generation}. \cref{tab:visual_grounding_single} shows that \modelname achieves comparable performance to the state-of-the-art methods. Our method achieve 0.97\% absolute improvement compared with the second best method on RefCOCO ``testB'' split without using object detection data for pre-training. Queries in ``testB'' split can refer to various visual concepts but only people in ``testA''. The improvement demonstrates that the introduction of universal layers can help model the visual concepts in the image.
\subsubsection{Language Only Tasks}

\begin{table}[t]
\centering
\caption{\textbf{Experimental results on the GLUE benchmark.} 
% For comparison, we list the performance of multimodal pre-trained models as well as the recent SOTA models that were pre-trained on natural language data only.
% All models are of \textit{Base} size, and they share a similar amount of parameters. 
% For models of \textit{Base} size, \ofa~outperforms all multimodal pretrained baseline models on the 77 tasks, and BERT on 66 tasks except CoLA. For models of \textit{Large} size, \ofa~performs competitively with \rm BERT_{Large}\rm BERT_{Large}. The reported results of the multimodal baselines are from \cite{iki2021effect} and the corresponding original papers. 
}
% \vskip 0.15in
\centering
% \begin{adjustbox}{max width=1.\textwidth}
\resizebox{1.0\linewidth}{!}{
\begin{tabular}{@{}lcccccc@{}}
\toprule
  Model
%   &CoLA
  &SST-2
  &RTE
  &MRPC
  &QQP
  &MNLI
  &QNLI
  \\
\midrule
\multicolumn{4}{l}{\textit{Multimodal Pretrained Baseline Models}}
%   &
  &
  &
  &
%   \\
%   BERT
%   &\textbf{54.6}
%   &92.5
%   &62.5
%   &81.9/87.6
%   &90.6/87.4
%   &84.2
%   &91.0
  \\
  VisualBERT~\citep{li2019visualbert}
%   &38.6
  &89.4
  &56.6
  &71.9
  &89.4
  &81.6
  &87.0
  \\
  UNITER~\citep{chen2020uniter}
%   &37.4
  &89.7
  &55.6
  &69.3
  &89.2
  &80.9
  &86.0
  \\
  VL-BERT~\citep{su2019vl}
%   &38.7
  &89.8
  &55.7
  &70.6
  &89.0
  &81.2
  &86.3
  \\
  VilBERT~\citep{lu2019vilbert}
%   &36.1
  &90.4
  &53.7
  &69.0
  &88.6
  &79.9
  &83.8
  \\
  LXMERT~\citep{tan2019lxmert}
%   &39.0
  &90.2
  &57.2
  &69.8
  &75.3
  &80.4
  &84.2
  \\
  Uni-Perceiver~\citep{uni-perceiver}
  &90.2
  &64.3
  &86.6
  &87.1
  &81.7
  &89.9
  \\
  SimVLM~\citep{wang2021simvlm}
%   &46.7
  &90.9
  &63.9
  &75.2
  &90.4
  &83.4
  &88.6
  \\
  FLAVA~\citep{Singh2021FLAVA}
%   &50.7
  &90.9
  &57.8
  &81.4
  &90.4
  &80.3
  &87.3
    \\
  UNIMO~\citep{li2020unimo}
%   &39.0
  &96.8
  &-
  &-
  &-
  &89.8
  &-
  \\
  OFA~\citep{Wang2022OFA}
%   &39.0
  &96.6 
  &\textbf{91.0} 
  &91.7
  &92.5
  &90.2
  &94.8
  \\
  \midrule
\multicolumn{5}{l}{\textit{Natural-Language-Pretrained SOTA Models}}
%   &
  &
  &
  \\
  BERT~\citep{devlin2018bert}
%   &60.6
  &93.2
  &70.4
  &88.0
  &91.3
  &86.6
  &92.3
  \\
  RoBERTa~\citep{liu2019roberta}
%   &68.0
  &96.4
  &86.6
  &90.9
  &92.2
  &90.2
  &93.9
  \\
  XLNet~\citep{Zhilin2019XLNet}
%   &69.0
  &\textbf{97.0}
  &85.9
  &90.8
  &92.3
  &90.8
  &94.9
  \\
  ELECTRA~\citep{electra}
%   &69.1
  &96.9
  &88.0
  &90.8
  &92.4
  &90.9
  &95.0
  \\
  DeBERTa~\citep{deberta}
%   &70.5
  &96.8
  &88.3
  &91.9
  &92.3
  &\textbf{91.1}
  &\textbf{95.3}
  \\
%   UniLM
%   &\textbf{61.1}
%   &94.5
%   &70.9
%   &-/-
%   &-/-
%   &\textbf{87.0}
%   &92.7
%   \\
\midrule
  \modelnamebase
%   &-
  & 93.5
  & 85.2
  & 87.3
  & 91.3
  & 87.6
  & 93.2
  \\
  \modelname
%   &-
  & 95.1
  & 88.0
  & 90.1
  & \textbf{92.7}
  & 90.2
  & 94.5
  \\
    \modelnamedeberta
%   &-
  & 96.2
  & 89.4
  & \textbf{92.1}
  & 92.6
  & 90.8
  & 94.8
  \\
\bottomrule
\end{tabular}
}
% \end{adjustbox}
\label{table:glue-results}
\vspace{-2ex}
\end{table}
\begin{table}[t]
\caption{\textbf{Experimental results on Gigaword abstractive summarization.} We report performance on the ROUGE evaluation.}
\centering
% \begin{adjustbox}{max width=1.\textwidth}
\resizebox{1.0\linewidth}{!}{
\begin{tabular}{lcccc}
\toprule
  \multirow{2}*{Model}
  & \multicolumn{3}{c}{Gigaword}
  \\
  & ROUGE-1 & ROUGE-2 & ROUGE-L
  \\
\midrule
  BERTSHARE~\citep{bertshare}
  & 38.13 & 19.81 & 35.62
  \\
  MASS~\citep{mass}
  & 38.73 & 19.71 & 35.96
  \\
  UniLM~\citep{unilm}
  & 38.45 & 19.45 & 35.75
  \\
  PEGASUS~\citep{pegasus}
  & 39.12 & 19.86 & 36.24
  \\ 
  ProphetNet~\citep{prophetnet}
  & 39.55 & 20.27 & 36.57
  \\ 
  UNIMO~\citep{li2020unimo}
  & 39.71 & 20.37 & 36.88
  \\ 
  OFA~\citep{Wang2022OFA}
  & \textbf{39.81} & 20.66 & \textbf{37.11}
  \\ 
\midrule
  \modelname
  & 39.65&\textbf{20.67}&36.89
  \\
\bottomrule
\end{tabular}
}
\vspace{-2ex}
% \end{adjustbox}
\label{table:nlg-results}
% \vskip -0.1in
\end{table}

\paragraph{Natural Language Understanding} We evaluate \modelname on 6 tasks of the GLUE benchmark ~\citep{Wang2018GLUEAM}  for natural language understanding. Table~\ref{table:glue-results} shows that \modelname achieves comparable performance to the state-of-the-art natural language and multimodal pretrained models including RoBERTa~\citep{liu2019roberta}, DeBERTa~\citep{deberta}. Our method with DeBERTa achieves improvement compared with DeBERTa~\citep{deberta} on three tasks, which also demonstrate the effectiveness of universal modules for modality collaboration.
\paragraph{Natural Language Generation} We evaluate \modelname on Gigaword abstractive summarization ~\citep{rush2015neural} for natural language generation. As shown in Table~\ref{table:nlg-results}\ref{table:nlg-results}, \modelname achieves the comparable result with  the state-of-the-art models.

\vspace{-2ex}
\subsubsection{Vision Only Tasks}

\begin{table}[t!]
\centering
\caption{\textbf{Comparison with the state-of-the-art on video action recognition under fine-tuning settings.}
    }
    \tablestyle{7pt}{1.1} 
    \def \w{15pt}
    \resizebox{0.99\linewidth}{!}{
    \begin{tabular}{l|cc|cc|cc}
        \shline
        ~ & \multicolumn{2}{c}{Kinetics 400} & \multicolumn{2}{c}{Kinetics 600} & \multicolumn{2}{c}{Kinetics 700} \\
        \cmidrule(lr){2-3} \cmidrule(lr){4-5} \cmidrule(lr){6-7}
        Method & Top-1 & Top-5 & Top-1 & Top-5 & Top-1 & Top-5  \\
        \shline
        TimeSformer-L \citep{bertasius2021timesformer} & 80.6 & 94.7 & 82.2 & 95.6 & - & - \\
        ViViT-H \citep{Anurag2021ViViT} & 84.8 & 95.8 & 85.8 & 96.5 & - & - \\
        VideoSwin-L \citep{Ze2022videoswin} & 84.9 & 96.7 & 86.1 & 97.3 & - & - \\
        OmniVL \citep{Wang2022OmniVLOF} & 79.1 & 94.5 & - & - & - & - \\
        TokenLearner \citep{Ryoo2021TokenLearnerWC} & 85.4 & 96.3 & 86.3 & 97.0 & - & - \\
        VATT \citep{akbari2021vatt} & 82.1 & 95.5 & 83.6 & 96.6 & - & - \\
        MoViNet \citep{Dan2021MoViNets} & 81.5 & - & 84.8 & - & 79.4 & - \\
        Florence \citep{yuan2021florence} & 86.5 & 97.3 & 87.8 & 97.8 & - & - \\
        CoVeR \citep{Zhang2021COVER} & 86.3 & 97.2 & 	87.9 & 97.8 & 78.5 & 94.2 \\
        \hline
        \modelnamebase & 83.6 & 96.0 & 86.7 & 97.2 & 74.6 & 91.2 \\
        \modelname  & \textbf{87.1} & \textbf{97.7} & \textbf{89.8} & \textbf{98.3} & \textbf{80.4} & \textbf{94.9} \\
        \shline
    \end{tabular}
    }
    \label{table:video-classification}
    \vspace{-2ex}
\end{table}
\begin{table}[t!]
\centering
    \caption{\textbf{Comparison with state-of-the-art methods in terms of robustness and generalization capability evaluation on ImageNet-1K variants.} We test the model on various ImageNet-1K validation set without any further fine-tuning. "Avg." indicates the average Top-1 accuracy on 6 different ImageNet-1K variants. "$\Delta_\downarrow$" stands for the gap between averaged Top-1 accuracy of 6 variants and the accuracy of original ImageNet-1K validation (the lower the better). 
    }
    \tablestyle{7pt}{1.1} 
    \def \w{15pt}
    \resizebox{0.99\linewidth}{!}{
    \begin{tabular}{l|cccccc|c|c}
        \shline
        Method & IN-1K & IN-V2 & IN-ReaL & IN-Adv. & IN-Ren. & IN-Ske. & Avg. & $\Delta_\downarrow$  \\
        \shline
        ConvNeXt \citep{Liu2022ConvNeXt} & 87.5 & 77.7 & 90.5 & 70.8 & 67.0 & 53.7 & 74.5 & 13.0 \\
        SwinV2-G \citep{Liu2021SwinTV} & 87.5 & 77.3 & 90.2 & 73.9 & 67.7 & 52.3 & 74.8 & 12.7 \\
        MAE \citep{He2021MAE} & 87.8 & 79.2 & 90.3 & 76.7 & 66.5 & 50.9 & 75.2 & 12.6 \\
        DeiT3 \citep{Touvron2022DeiT3} & 87.7 & 79.1 & 90.2 & 79.2 & 70.6 & 54.9 & 77.0 & 10.7 \\
        Eff-L2-NS \citep{Tan2019EfficientNetRM} & 88.4 & \textbf{80.5} & \textbf{90.6} & \textbf{84.8} & 74.7 & 47.6 & \textbf{77.8} & \textbf{10.6} \\
        OFA \citep{Wang2022OFA} &  85.6 & - & - & - & - & - & - & - \\
        % BEiTv2 \citep{} & 88.4 & 80.1 & 90.3 & 76.2 & 76.4 & 58.3 & 78.3 & 10.1 \\
        \hline
        \modelname & \textbf{88.5} & 78.1 & 89.5 & 73.2 & \textbf{75.6} & \textbf{61.2} & 77.7 & 10.7 \\
        \shline
    \end{tabular}
    }
    \label{table:image-classification}
    \vspace{-2ex}
\end{table}

\begin{table}[t!]
\centering
\caption{\textbf{Evaluation of the proposed instructional-based learning on downstream tasks.} For retrieval task, we report the average of Recall@1, Recall@5, and Recall@10. For QA and caption task, Top-1 Accuarcy and CIDEr are reported.
    }
    \tablestyle{7pt}{1.1} 
    \def \w{15pt}
    \resizebox{0.95\linewidth}{!}{
    \begin{tabular}{c|ccc}
        \shline
        Instruction & MSRVTT-Ret. & MSVD-QA & MSRVTT-Cap. \\
        \shline
        ~ & 72.8 & 54.1 & 71.8 \\
        \hline
        $\checkmark$ & \textbf{73.5} \color{red}{(+0.7)} & \textbf{55.3} \color{red}{(+1.2)} & \textbf{72.4} \color{red}{(+0.6)} \\
        \shline
    \end{tabular}
    }
    \vspace{-2ex}
    \label{table:instruction-ablation}
\end{table}

\begin{table}[t!]
\centering
\caption{\textbf{Evaluation of different temporal modeling modules in the dual-vision encoder module.} For retrieval task, we report the average of Recall@1, Recall@5, and Recall@10. For QA and caption task, Top-1 Accuarcy and CIDEr are reported.
    }
    \tablestyle{7pt}{1.1} 
    \def \w{15pt}
    \resizebox{0.95\linewidth}{!}{
    \begin{tabular}{c|ccc}
        \shline
        Temporal Module & MSRVTT-Ret. & MSVD-QA & MSRVTT-Cap. \\
        \shline
        Temporal Self-Attention & 70.3 & 55.1 & 71.1 \\ \hline
        Temporal Convolution & 71.4 \color{red}{(+1.1)} & 55.0 \color{green}{(-0.1)} & 71.7 \color{red}{(+0.6)} \\ \hline
        Local Temporal Modeling & \textbf{73.5} \color{red}{(+3.2)} & \textbf{55.3} \color{red}{(+0.2)} & \textbf{72.4} \color{red}{(+1.3)} \\
        \shline
    \end{tabular}
    }
    \vspace{-2ex}
    \label{table:temporal-ablation}
\end{table}

\begin{table*}[t!]
\centering
\caption{\textbf{Evaluation of the impact of universal layer} in terms of boosting vision task's performance.
    }
    \tablestyle{7pt}{1.1} 
    \def \w{15pt}
    \resizebox{0.95\linewidth}{!}{
    \begin{tabular}{l|ccccccc|c}
        \shline
        Method & ImageNet & CIFAR10 & CIFAR100 & Cars & DTD & SUN & Food101 & Average \\
        \shline
        CLIP-ViT-L/14 & 86.2 & 98.6 & 92.2 & 91.6 & 81.9 & 80.7 & 94.4 & 89.4  \\
        \hline
        +Universal Layers & 86.6 \color{red}{(+0.4)} & 99.3 \color{red}{(+0.7)} & 93.1 \color{red}{(+0.9)} & 94.4 \color{red}{(+2.8)} & 85.1 \color{red}{(+3.2)} & 80.4 \color{green}{(-0.4)} & 95.4 \color{red}{(+1.0)} & \textbf{90.6} \color{red}{(+1.2)} \\
        \shline
    \end{tabular}
    }
    \vspace{-2ex}
    \label{table:clip-ablation}
\end{table*}

\begin{table}[t]
\centering
\caption{\textbf{Evaluation of the impact of the universal layer} in terms of boosting language and vision-language task's performance.
}
% \vskip 0.15in

% \begin{adjustbox}{max width=1.\textwidth}
\resizebox{0.95\linewidth}{!}{
\begin{tabular}{@{}l|cccccc|c@{}}
\toprule
  Model
%   &CoLA
  &SST-2
  &RTE
  &MRPC
  &QQP
  &MNLI
  &QNLI
  &VQA test-dev
  \\
\midrule
  $\text{BERT}_{base}$
%   &-
  & 91.7
  & 71.4
  & 86.3
  & 90.8
  & 84.3
  & 89.3
  & 78.6
  \\
  +Joint Training
%   &-
  & 92.5
  & 82.3
  & 86.6
  & 90.6
  & 86.2
  & 92.1
  & 78.9
  \\
+Universal Layers
%   &-
  & \textbf{93.5}
  & \textbf{85.2}
  & \textbf{87.3}
  & \textbf{91.3}
  & \textbf{87.6}
  & \textbf{93.2}
  & \textbf{79.3}
  \\
\bottomrule
\end{tabular}
}
% \end{adjustbox}
\label{table:glue-vqa-ablation}
\end{table}

\paragraph{Video Action Recognition}
Video action recognition is the most representative for video understanding since it requires the model to understand the spatio-temporal cues revealed in the video. Table \ref{table:video-classification} summarizes the performance of different approaches on Kinetics 400, Kinetics 600, and Kinetics 700 datasets. Our \modelname surpasses the most of SoTA methods. For example, comapred with Florence pre-trained on 900M vision-text pairs, \modelname improves the Top-1 accuracy by 1.9\% on Kinetics 600 and 0.6\% on Kinetics 400. Meanwhile, we can notice that the performance of \modelname is better than OmniVL with similar amount of pre-training data, which shows the effectiveness of the dual-vision encoder module for video representation learning.

\paragraph{Image Classification}
We further evaluate the performance of \modelname in terms of image classification on ImageNet-1K. As we can see in the Table \ref{table:image-classification}, We can see that \modelname achieves comparable results or even surpass the SoTA methods on ImageNet-1K without using the ImageNet data for pre-training. Besides, to effectively evaluate the robustness and generalization ability of \modelname, we perform the evaluation on 5 ImageNet variants (i.e. IN-V2, IN-Real., IN-Adversarial, IN-Rendition, and IN-Sketch). Following standard evaluation procedure \citep{Fang2022EVA}, all these models are first fine-tuned on the original ImageNet-1K training set and directly tested on the 6 variants without further fine-tuning. As shown in Table \ref{table:image-classification}, \modelname not only achieves the highest accuracy on ImageNet-1K validation set but also obtains the relative small gap (i.e., $\Delta_\downarrow$), which reflects the excellent robustness and generalization capability of \modelname with the help of the universal layer module by learning language-shared representation.

\subsection{Discussion}
\paragraph{Impact of Instruction-based Learning}
The instructional-based learning is able to distinguish different types of tasks with specific instructions. Table \ref{table:instruction-ablation} demonstrates the effectiveness of instructional-based learning. In the table, we can observe that the instructional-based learning improves the performance of retrieval and question answering by at least 0.7\% and 1.2\% in Average Recall and accuracy respectively. With the help of instructional-based learning, \modelname is capable of utilizing the different modules when different instructions are used to boost the performance.

\paragraph{Impact of Local Temporal Modeling Module}
To validate the effectiveness of our proposed local temporal modeling module in the dual-vision encoder, we conduct experiments with the different temporal modeling structures. Specially, we have tried out the temporal self-attention and temporal convolution for comparison. The results are summarized in Table \ref{table:temporal-ablation}. We can notice that the local temporal modeling module outperforms temporal self-attention module by introducing modeling temporal locality. Meanwhile, with the help of the multi-group fusion mechanism, the local temporal modeling module can learn the diverse temporal representations in distinctive representation subspaces while the temporal convolution is restricted in the same temporal representation spaces, thus leading to the better performance.

\paragraph{Impact of Universal Layer}
To validate the effectiveness of our proposed universal layer module, we ablation this module for all uni-modal and multi-modal tasks. 
% In Table \ref{table:clip-ablation}, we can observe that the shared universal layer can boost the performance. 
As shown in Table \ref{table:clip-ablation} and Table \ref{table:glue-vqa-ablation}, we set Row 1/2/2 as the baseline of the vision/language/vision-language task in this experiment, respectively.
We can find that compared with the baseline the shared universal layer is beneficial for all modality tasks by encouraging collaboration between modalities.

\begin{figure}
    \centering
    \includegraphics[width=0.89\linewidth]{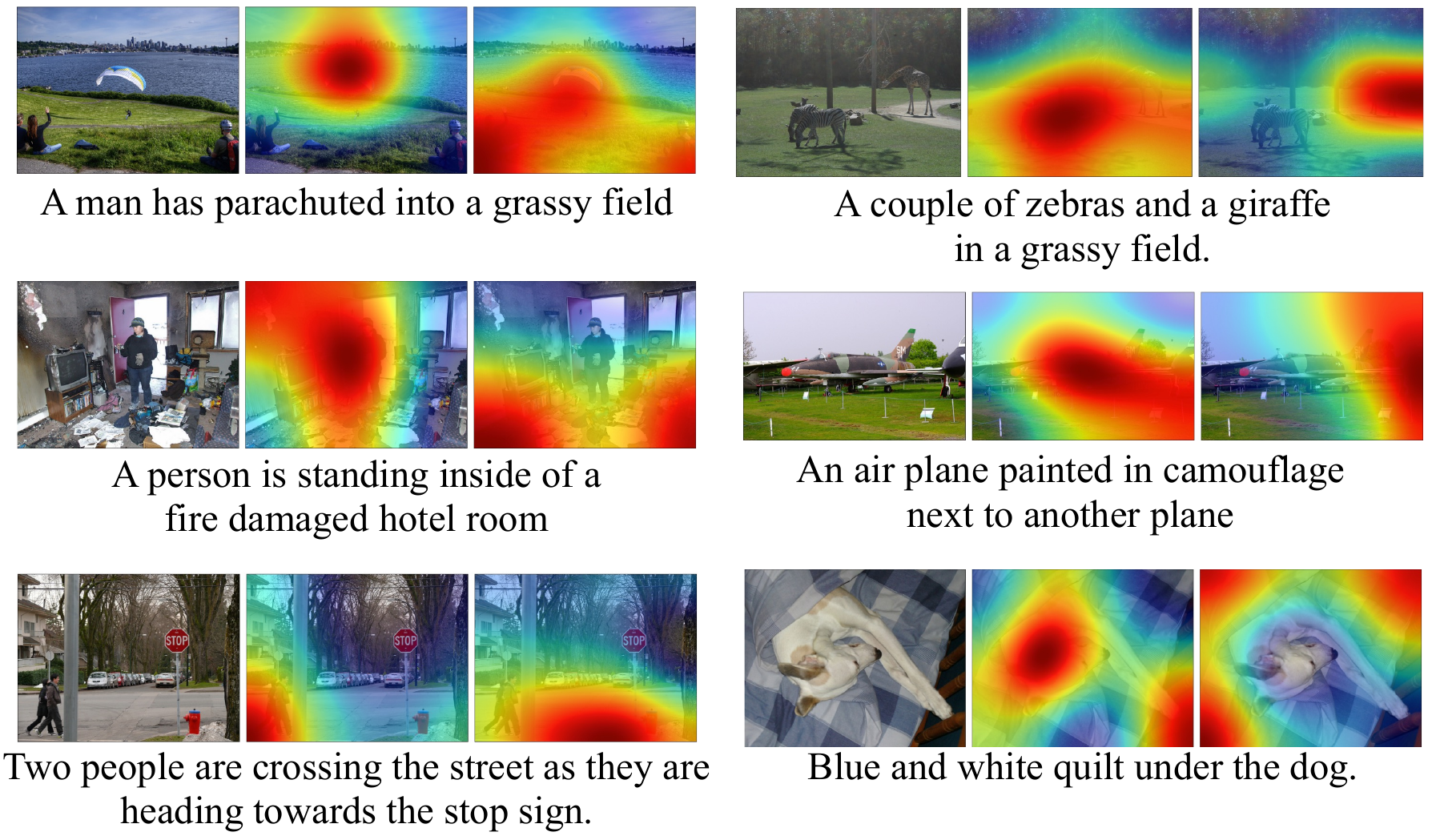}
\caption{Grad-CAM visualizations for latent queries in the universal layers.}
\vspace{-2ex}
\label{fig:latent_visu}
\end{figure}
In \cref{fig:latent_visu}, we visualize the Grad-CAM on the cross-attention map in the first universal layer. For each sample, we present two cross-attention maps that attend to different visual concepts. The results show that the universal layer can encourage modality collaboration and modality entanglement between visual patch features and language features by attending the areas of various visual concepts in the image.

\paragraph{Universal Layer for Modality Collaboration}
\begin{figure}
    \centering
    \includegraphics[width=0.99\linewidth]{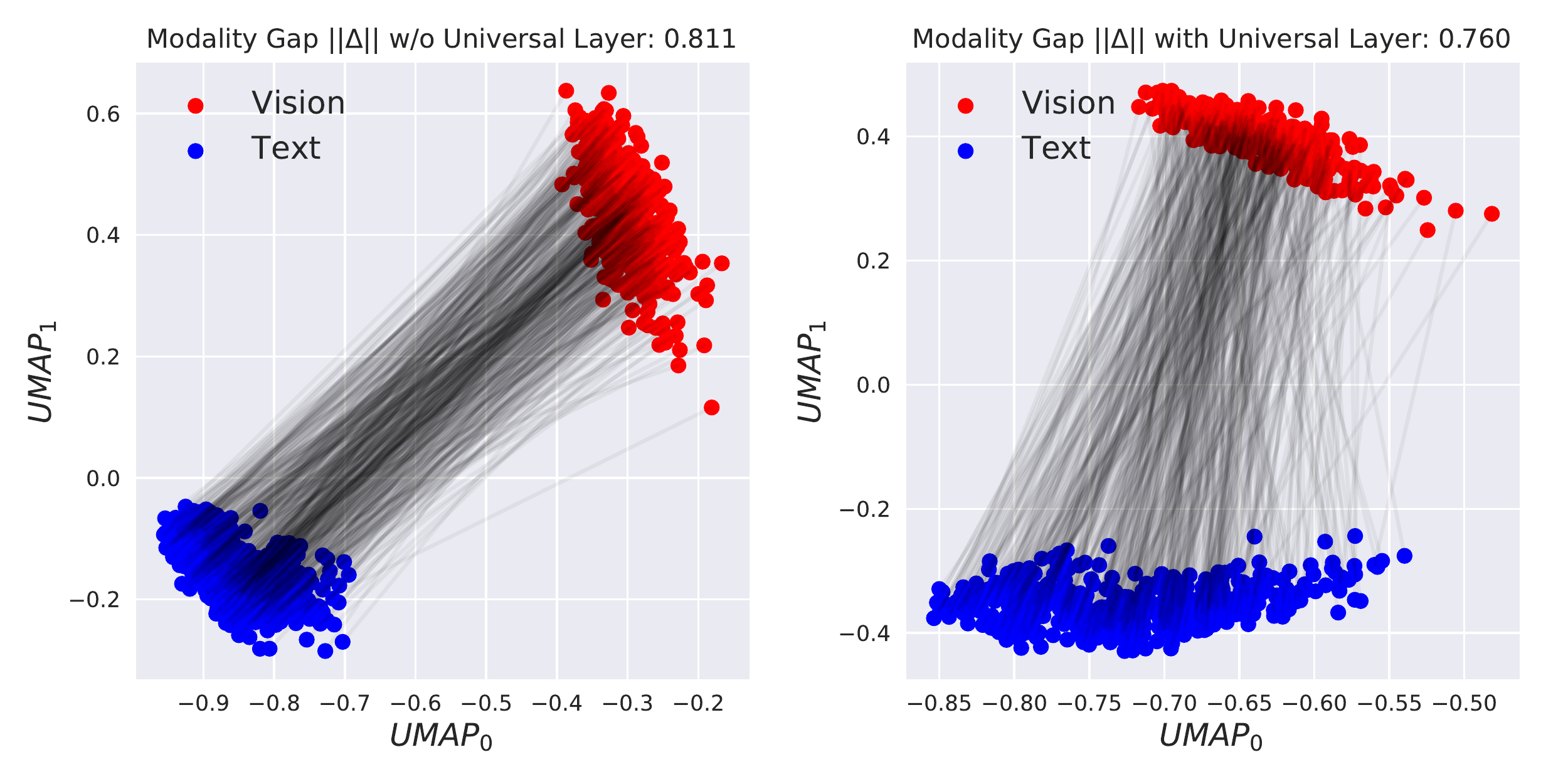}
    \vspace{-3ex}
\caption{The UMAP visualization of generated vision and language embeddings from pre-trained \modelname. The black lines refer to vision-language pairs.}
\vspace{-2ex}
\label{fig:modality-gap}
\end{figure}
Here we investigate the influence of universal layer in terms of modality collaboration. We randomly sample some vision-language pairs, and sketch the UMAP visualization of the generated embeddings from pre-trained \modelname in the Figure \ref{fig:modality-gap}. We can observe that with the help of universal layer, the distance between vision and text samples are more closer instead of solely two concentrated clusters. Besides, we quantitatively compute the modality gap $\|\Delta\|$ \citep{liang2022mind}, where the $\Delta$ is the difference between the center of vision embeddings and text embeddings. It can be observed that the model with universal layer would encourage the collaboration between vision and language modalities thus yielding lower modality gap compared to the model without universal layer.

\vspace{-2ex}
\section{Conclusion}
% In the unusual situation where you want a paper to appear in the
% references without citing it in the main text, use \nocite
%\nocite{langley00}
This paper presents \modelname, a new unified paradigm with modularized design for building multi-modal foundation models. \modelname introduces a module-based network design that shares common universal modules for modality collaboration and disentangles modality-specific modules to address the problem of modality entanglement. Experimental results show that the new unified paradigm of \modelname can achieve strong performances on a broad range of over 30 tasks across the text, image and video modalities. It is also easy to extend \modelname to more tasks by selecting and adding modules.

% \clearpage

\bibliography{mplug}
\bibliographystyle{icml2023}

%%%%%%%%%%%%%%%%%%%%%%%%%%%%%%%%%%%%%%%%%%%%%%%%%%%%%%%%%%%%%%%%%%%%%%%%%%%%%%%
%%%%%%%%%%%%%%%%%%%%%%%%%%%%%%%%%%%%%%%%%%%%%%%%%%%%%%%%%%%%%%%%%%%%%%%%%%%%%%%
% APPENDIX
%%%%%%%%%%%%%%%%%%%%%%%%%%%%%%%%%%%%%%%%%%%%%%%%%%%%%%%%%%%%%%%%%%%%%%%%%%%%%%%
%%%%%%%%%%%%%%%%%%%%%%%%%%%%%%%%%%%%%%%%%%%%%%%%%%%%%%%%%%%%%%%%%%%%%%%%%%%%%%%
\newpage
\appendix
\onecolumn

\section{More Results}

\subsection{Detection and Segmentation}
\begin{table}[t!]
\centering
    \caption{\textbf{System-level comparisons with the state-of-the-art results on COCO dataset for object detection and instance segmentation.} We report the standard boudning box AP (AP$_{box}$) and mask AP (AP$_{mask}$). The detector are Cascade Mask R-CNN (Cascade), Dynamic Head (DyHead), Hybrid Task Cascade (HTC), and its extension (HTC++).
    }
    \tablestyle{7pt}{1.1} 
    \def \w{15pt}
    % \resizebox{1.0\linewidth}{!}{
    % \begin{tabular}{lc|cc}
    %     \shline
    %     Method & Detector & AP$_{box}$ & AP$_{mask}$ \\
    %     \shline 
    %     Copy-Paste \citep{Golnaz2021copypaste} & Cascade & 57.0 & 48.9 \\
    %     Swin-L \citep{liu2021Swin} & HTC++ & 58.0 & 50.4 \\
    %     MViTv2-H \citep{Li2021MViTv2IM} & Cascade & 58.4 & 50.1 \\
    %     CBNetV2 \citep{Liang2021CBNetAC} & HTC & 59.6 & 51.8 \\
    %     Soft-Teacher \citep{Xu2021softteacher} & Cascade & 60.7 & 52.5 \\
    %     GLIP \citep{Li2021GLIP} & DyHead & 60.8 & - \\
    %     SwinV2-L \citep{Liu2021SwinTV} & HTC++ & 60.2 &  52.1 \\
    %     Florence \citep{yuan2021florence} & DyHead & 62.0 & - \\
    %     BEiT-3 \citep{Wang2022BEITv3} & Cascade & 63.7 & 54.8 \\
    %     \hline 
    %     \modelname & Cascade & 46.9 & 40.6 \\
    %     \shline
    % \end{tabular}
    \begin{tabular}{lc|cc}
\hline
Method                                 & Detector & AP$_{box}$ & AP$_{mask}$ \\ \hline
Mask R-CNN \citep{He2017MaskR}         & -        & 46.3       & 40.1        \\
DETR \citep{Nicolas2020DETR}           & -        & 44.9       & 33.0        \\
Pix2seq \citep{chen2021pix2seq}        & -        & 45.0       & -           \\
Copy-Paste \citep{Golnaz2021copypaste} & Cascade  & 57.0       & 48.9        \\
Swin-L \citep{liu2021Swin}             & HTC++    & 58.0       & 50.4        \\
%MViTv2-H \citep{Li2021MViTv2IM}        & Cascade  & 58.4       & 50.1        \\
CBNetV2 \citep{Liang2021CBNetAC}       & HTC      & 59.6       & 51.8        \\
%Soft-Teacher \citep{Xu2021softteacher} & Cascade  & 60.7       & 52.5        \\
GLIP \citep{Li2021GLIP}                & DyHead   & 60.8       & -           \\
SwinV2-L \citep{Liu2021SwinTV}         & HTC++    & 60.2       & \textbf{52.1}        \\
Florence \citep{yuan2021florence}      & DyHead   & \textbf{62.0}       & -           \\ \hline
\modelname                             & Cascade  & 46.9       & 40.6        \\ \hline
\end{tabular}
    % }
    \label{table:coco-detection}
\end{table}

We evaluate the object detection and instance segmentation performance of \modelname on COCO dataset \citep{lin2014microsoft}, which is widely used for object-level detection and segmentation with 80 common categories. Table \ref{table:coco-detection} reports the results on COCO dataset. We observe that \modelname outperform typical state-of-the-art resnet-based detection methods (e.g., DETR~\citep{Nicolas2020DETR} and Pix2seq~\citep{chen2021pix2seq}). There is a performance gap between foundation model optimized for computer vision (e.g., Florence \citep{yuan2021florence} and Swin-Transformer \citep{liu2021Swin}) and \modelname. Note that \modelname 
does not pre-trained with vision only task and data. Lower performance than models pre-trained on ImageNet is to be expected.

% \subsection{Impact of Local Temporal Modeling Module}
% \input{tables/local_temporal_ablation}
% To validate the effectiveness of our proposed local temporal modeling module in the dual-vision encoder, we conduct experiments with the different temporal modeling structures. Specially, we have tried out the temporal self-attention and temporal convolution for comparison. The results are summarized in Table \ref{table:temporal-ablation}. We can notice that the local temporal modeling module outperforms temporal self-attention module by introducing modeling temporal locality. Meanwhile, with the help of the multi-group fusion mechanism, the local temporal modeling module can learn the diverse temporal representations in distinctive representation subspaces while the temporal convolution is restricted in the same temporal representation spaces, thus leading to the better performance.

\subsection{Zero-Shot Transferability}
\paragraph{Text-to-Video Retrieval}
\begin{table*}[t!]
\centering
    \caption{\textbf{Zero-shot evaluation on text-to-video retrieval.} All results are reported on R@1/R@5/R@10.
    }
    \tablestyle{7pt}{1.1} 
    \def \w{15pt}
    \resizebox{0.8\linewidth}{!}{
    \begin{tabular}{ll|ccc|ccc|ccc}
        % \toprule
        \shline
        ~ & ~ & \multicolumn{3}{c}{MSRVTT}& \multicolumn{3}{c}{DiDeMo} & \multicolumn{3}{c}{LSMDC} \\
        \cmidrule(lr){3-5} \cmidrule(lr){6-8} \cmidrule(lr){9-11}
        Method & \# PT Data & R@1 & R@5 & R@10 & R@1 & R@5 & R@10 & R@1 & R@5 & R@10 \\
        \shline
        Frozen~\citep{bain2021frozen} & 5M  & 18.7 & 39.5 & 51.6 & 21.1 & 46.0 & 56.2 & 9.3 & 22.0 & 30.1 \\
        ALPRO~\citep{li2021align_prompt} & 5M & 24.1 & 44.7 & 55.4 & 23.8 & 47.3 & 57.9 & - & - & -  \\
        
        Singularity~\citep{lei2022singularity} & 5M & 28.4 & 50.2 & 59.5 & 36.9 & 61.6 & 69.3 & - & - & - \\
        \hline
        VIOLET~\citep{fu2021violet} & 183M & 25.9 & 49.5 & 59.7 & 23.5 & 49.8 & 59.8 & - & - & - \\
        Florence~\citep{yuan2021florence} & 900M & 37.6 & 63.8 & 72.6 & - & - & - & - & - & - \\
        mPLUG \citep{Li2022mPLUGEA} & 14M & 38.1 & 59.2 & 68.2 & - & - & - & - & - & - \\
        HiTeA \citep{Ye2022HiTeAHT} & 17M & 34.4 & 60.0 & 69.9 & 43.2 & 69.3 & 79.0 & 18.3 & 36.7 & 44.2 \\
        OmniVL \citep{Wang2022OmniVLOF} & 18M & 42.0 & 63.0 & 73.0 & 40.6 & 64.6 & 74.3 & - & - & - \\
        \hline
        \modelname & 17M & \textbf{47.1} & \textbf{69.7} & \textbf{79.0} & \textbf{45.7} & \textbf{71.1} & \textbf{79.2} & \textbf{24.1} & \textbf{43.8} & \textbf{52.0} \\
        \shline
        % \bottomrule
    \end{tabular}
    }
    \vspace{-1ex}
    \label{table:videoretrieval-zeroshot}
    \vspace{-2ex}
\end{table*}
For testing the transferability of pre-trained \modelname, we conduct the zero-shot evaluation on Text-to-Video Retrieval and the results are summarized in Table \ref{table:videoretrieval-zeroshot}. We can find that \modelname obtains SoTA results on both MSRVTT, DiDeMo and LSMDC datasets, and outperforms previous methods by a large margin, such as 5.1 point of R@1 on the MSRVTT dataset. The results prove that our \modelname has excellent zero-shot transferability.

\paragraph{Video Question Answering}

\begin{table}[t!]
\centering
\caption{\textbf{Zero-shot evaluation on video question answering.} Accuracy is reported.}
    \tablestyle{7pt}{1.1} 
    \def \w{15pt}
    % \resizebox{\linewidth}{!}{
    \begin{tabular}{ll|cc}
        % \toprule
        \shline
        % ~ & ~ & \multicolumn{4}{c}{Val} \\
        % \cmidrule(lr){3-6}
        Method & \# PT Data & MSRVTT-QA & MSVD-QA \\
        \hline
        Just Ask~\citep{yang2021justask} & 69M & 2.9 & 7.5 \\
        LAVENDER~\citep{li2022lavender} & 5M & 4.5 & 11.6 \\
        MERLOT Reserve~\citep{zellers2021merlot} & 1B & 5.8 & - \\
        FrozenBiLM~\citep{yang2022frozenbilm} & 10M & 6.4 & 11.7 \\
        BLIP~\citep{li2022blip} & 129M  &  19.2 & 35.2 \\
        mPLUG~\citep{Li2022mPLUGEA} & 400M & 21.1  & 37.2 \\
        HiTeA~\citep{Ye2022HiTeAHT} & 5M & 21.7 & 37.4 \\
        \hline
        \modelname & 17M & \textbf{43.8} & \textbf{55.3} \\
        \shline
        % \bottomrule
    \end{tabular}
    % }
    \label{table:videoqa-zeroshot}
\end{table}
We testing the transferability of pre-trained \modelname on Video QA and the results are summarized in the Table \ref{table:videoqa-zeroshot}. It can be observed that \modelname achieves the best zero-shot performance on both MSRVTT-QA and MSVD-QA datasets, which demonstrates the strong zero-shot transferability of our model under the help of universal module and instructional-based learning.

\subsection{Visual Grounding}
\begin{figure}
    \centering
    \includegraphics[width=0.8\linewidth]{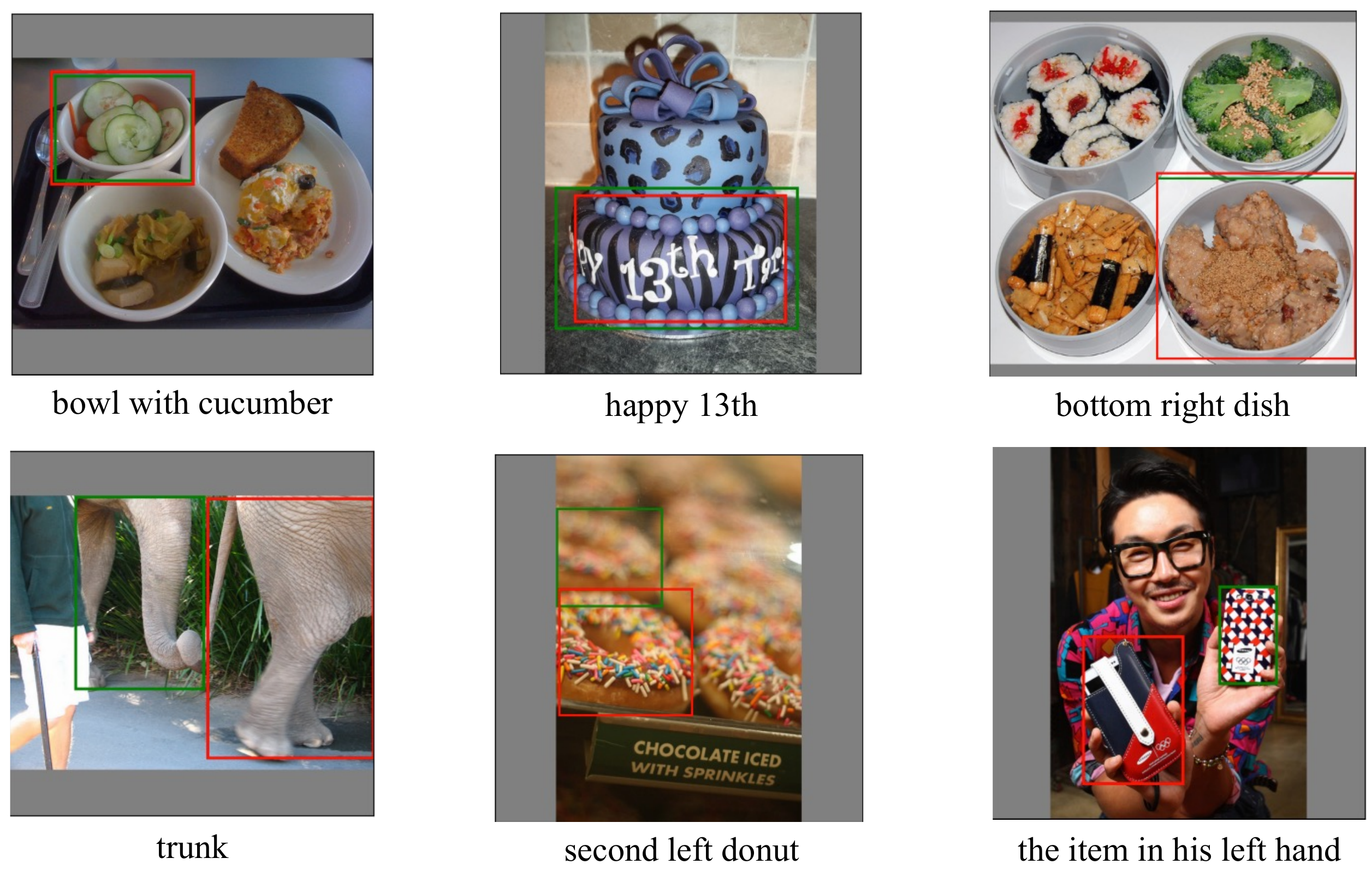}
\caption{The visualization of visual grounding task. Green denotes the ground-truth box and red denotes the predicted bounding box.}
\label{fig:vg_visu}
\end{figure}
We visualize several cases of visual grounding task in \Cref{fig:vg_visu}. The first row shows that our \modelname can understand various visual concepts and their relationships. It also can make fine-grained alignment between vision and language. The second row presents several failure cases. In the first sample, ``trunk'' is an ambiguous which result in a incorrect prediction. In the second sample, \modelname fail to recognize the blurred ``donut''. In the thrid sample, \modelname does not realize the left and right are reversed in a mirror and predict the ``left'' item. 
% 如果需要具体分析可以用这一版 如果觉得需要很多图堆在visulalize实验里可以用mplug 1.0那版 
\section{Implementation Details}

\subsection{Pre-training}
Our models are implemented in the PyTorch framework \citep{Paszke2019PyTorchAI}. In detail, we instantiate the text encoder with BERT \citep{devlin2018bert} model pre-traiend on Wikipedia and Bookcorpus \citep{Zhu2015BookCorpus}. The visual encoder is initialized from CLIP-ViT \citep{radford2021learning} pre-trained on 400M noisy image-text pairs. For the base size of model namely \modelnamebase, we use the ViT-B/16 for vision encoder and BERT-Base \citep{devlin2018bert} as the text encoder as well as the text decoder. For \modelname, we scale up the vision and text encoders with ViT-L/14 \citep{dosovitskiy2020image} and BERT-Large \citep{devlin2018bert} respectively. $C=768$ and $C=1024$ for \modelnamebase and \modelname. We set $S=2$ for universal layers for the good empirical performance, and choose $G=C$ for multi-group mechanism in the local temporal modeling module empirically. The number of layers for fusion module is set to 3 for \modelnamebase and 6 for \modelname, while the number of shared decoder layer is set to 12 for both \modelnamebase and \modelname. We pre-train the model for 30 epochs with the total batch size of 1024 on 8 NVIDIA A100 GPUs for \modelnamebase and batch size of 512 on 16 NVIDIA A100 GPUs. We use AdamW \citep{Loshchilov2019DecoupledWD} optimizer with the weight decay factor 0.02 and betas (0.9, 0.98) for stabilizing the learning. The learning rate is firstly warmed up to $lr_{max}$ in the first 5000 iterations then decays following the cosine annealing schedule. $lr_{max}$ is set to 1e-4 for \modelnamebase and 5e-5 for \modelname. During the pre-training, we randomly crop the images and video frames into $224\times 224$ resolution and sparsely sample 4 frames for each video while preserving their order in-between. For vision-text contrastive learning, the queue size and the momentum coefficient are set to 65,536 and 0.995 respectively.

\subsection{Downstream Tasks}
\subsubsection{Vision Only Tasks}
\paragraph{Video Action Recognition}
We first train \modelname on the Kinetics-710 dataset for 40 epochs which is the combination of Kinetics-400, Kinetics-600 and Kinetics-700 by removing the videos represented in the validation and test sets. Specially, the base learning rate is set to 1e-5 for \modelnamebase and 5e-6 for \modelname with batch size 256 and 128 respectively. Then fine-tuning on Kinetics-400, Kinetics-600, and Kinetics-700 individually for 5 epochs with same learning rate and batch size.

\paragraph{Image Classification}
We finetune \modelname for 30 epochs with the learning rate of 6e-5 and a batch size of 4096. 
we use the RandomCrop, HorizontalFlip, RandAug and RandErase transformations for data augmentation.

\paragraph{Object Detection and Segmentation}
We keep the same setting as EVA~\citep{Fang2022EVAET} to train \modelname on object detection and segmentation tasks. The different is that we do not pre-train \modelname on Object365~\citep{Shao2019Objects365AL} before fine-tuning on MSCOCO.

\subsubsection{Language Only Tasks}

\paragraph{Natural Language Understanding}
Following~\citep{Wang2022OFA}, we select the best hyperparameters in a suitable range for fine-tuning.
We tune the training epochs among {5, 7, 10}, learning rate among {3e-5, 5e-5, 6e-5, 7e-5, 1e-4}, batch size among {32, 64, 128}. We report the best performance on the development set for each task.

\paragraph{Natural Language Generation}
Following~\citep{Wang2022OFA}, we finetune \modelname for 50,000 steps with a learning rate of 3e-5 and a batch size of 256. 
During reference, we beam size with 5 and max generation length with 512.

\subsubsection{Video-Text Multi-modal Tasks}

For all video-language downstream tasks, we resize video frames to 224 $\times$ 224. 
During fine-tuning, we randomly sample 12 frames for text-to-video size video frames, 16 frames for video question answering and video captions. 
We perform uniform sampling during inference. 
We use RandomCrop with minimum ratio 0.5 and HorizontalFlip with 0.5 probability for data augmentation.

\paragraph{Text-to-Video Retrieval}
We train \modelnamebase and \modelname on the training set of MSRVTT/DiDeMo/LSMDC for 10 epochs with a learning rate of 2e-5 and batch size of 192.

\paragraph{Video Question Answering}
We train \modelnamebase and \modelname on the training set of MSRVTT-QA/MSVD-QA/TGIF-FrameQA for 10 epochs with a learning rate of 2e-5 and batch size of 128.

\paragraph{Video Captioning}
For the video caption task, we use a prefix prompt "What does the video describe?" to improve the quality of generated captions.
We set the same training parameters for both the MSRVTT and MSVD datasets. Specifically, we fine-tune \modelnamebase and \modelname with cross-entropy loss on the training set for 10 epochs with a learning rate of 2e-5 and a batch size of 128.
Then, we perform CIDEr optimization for extra 5 epochs with a learning rate of 1e-6 and a batch size of 16.
Finally, we evaluate the test set with a beam size of 5 and max generation length of 25.

\subsubsection{Image-Text Multi-modal Tasks}

We resize image frames to 336/576/384/336 for the retrieval/vqa/captioning/grounding tasks.
We use ResizedCrop with a minimum ratio of 0.5 and HorizontalFlip with 0.5 probability for data augmentation.
We perform center crop during inference. 

\paragraph{Image-Text Retrieval}

% We train \modelnamebase on the training set of MSCOCO/Flickr30K for xx/xx epochs with a learning rate of xx/xx and batch size of xx/xx.
We train \modelname on the training set of MSCOCO/Flickr30K for 8 epochs with a learning rate of 1e-5 and batch size of 512.

\paragraph{Visual Question Answering}
We train \modelnamebase on the VQA dataset for 8 epochs with a learning rate of 3e-5 and batch size of 512.

\paragraph{Image Captioning}

For the image caption task, we use a prefix prompt "What does the image describe ?" to improve the quality of generated captions.
we first fine-tune \modelname with cross-entropy loss on COCO training set for 5 epochs with a learning rate of 1e-5 and a batch size of 256. Then we evaluate on the COCO Caption Karpathy validation split and reuse it to predict the Nocaps validation set directly. During inference, we use beam search with a beam size of 5 and set the maximum generation length as 25.

\paragraph{Visual Grounding}
We first train the model with RefCOCO series datasets with a learning rate of 2e-5 for 120 epochs. Then we continue fine-tuning the model on each dataset with a learning rate of 2e-6 epochs for 30 epochs. We limit the query length to 20/40 for RefCOCO and RefCOCOg, respectively.

\subsection{Dataset Description}
\paragraph{Text-to-Video Retrieval} We evaluate \modelname on three popular
text-to-video retrieval datasets including MSRVTT~\citep{xu2016msrvtt},
DiDeMo~\citep{anne2017didemo}, and LSMDC~\citep{rohrbach2015lsmdc}.
% the cites refer to qinghao ~ HiTEA
\begin{itemize}
    \item \textbf{MSRVTT} consists of 10K YouTube sourced videos with 200K text descriptions. Following~\citep{li2022lavender, luo2022clip4clip, huang2022clover}, the dataset is divided into 9K and 1K videos for training and testing.
    \item \textbf{DiDeMo} consists of 10K videos from Flickr and each video with 4 descriptions. Following~\citep{li2022alpro, ma2022xclip, li2022lavender}, we concatenate all descriptions of a video as a paragraph, and evaluate the paragraph-to-video retrieval performance. The dataset is separated into 8K for training, 1K for validation and 1K videos for test.
    \item \textbf{LSMDC} consists of 118,081 video clips from 202 movies. Following the standard splits from~\citep{rohrbach2015lsmdc}, the dataset is divided into 101K and 1K videos for training and testing.

\end{itemize}

\paragraph{Video Question Answering}
We evaluate \modelname on three popular
video question answering datasets including MSRVTT-QA~\citep{xu2017msrvttqa} MSVD-QA~\citep{xu2017msrvttqa}, and TGIF-FrameQA~\citep{jang2017tgif}.
% the cites refer to qinghao ~ HiTEA

\begin{itemize}
    \item \textbf{MSRVTT-QA} is based on the MSRVTT dataset~\citep{xu2016msrvtt}. The QA pairs are automatically generated by from the descriptions. This benchmark composed of 243K open-ended questions over 10K videos.
    \item \textbf{MSVD-QA} is based on the MSVD datasets~\citep{chen2011msvd} with automatically generated QA pairs. It consists 2K videos with 47K questions.
    \item \textbf{TGIF-FrameQA} collects the answerable with just a single frame in the video, and is divided into training set with 35K questions and test set with 14K questions.
\end{itemize}

\paragraph{Video Captioning}
We use MSRVTT~\citep{xu2016msrvtt} and MSVD~\citep{chen2011msvd} for video captioning evaluation.

\begin{itemize}
    \item \textbf{MSRVTT} is composed of 10K videos with 20 captions per video as described above. We take the same data split as text-to-video retrieval task. 
    \item \textbf{MSVD} contains 1970 YouTube short video clips. 
    Following the standard splits from~\citep{lin2022swinbert, li2022lavender}, we separate the dataset into 1,200 train, 100 validation and 670 test videos.
    % the cite refers to ORG-TRL
\end{itemize}

\paragraph{Visual Question Answer}
We evaluate our method on the VQA 2.0 dataset~\citep{vqa}.

\begin{itemize}
    \item \textbf{VQA 2.0} is a dataset containing open-ended questions about images and at least 3 questions (5.4 questions on average) per image. It contains 83k/41k/81k images for training/validation/test.
\end{itemize}

\paragraph{Image-Text Retrieval}
Two popular image-text retrieval benchmarks, COCO~\citep{lin2014microsoft} and Flickr30K~\citep{plummer2015flickr30k} are used to evaluate the model. We adopt the widely-used Karpathy split~\citep{karpathy2015deep} for both COCO and Flickr30K.
% cites refer to chaoya~ patch merge
\begin{itemize}
    \item \textbf{COCO} has over 330k images and 5 independent human generated captions are be provided for each image. It contains 113k/5k/5k images for training/validation/testing.

    \item \textbf{Flickr30K} contains 31k images from Flickr, each image with 5 human annotated sentences. It contains 29k/1k/1k images for training/validation/testing.
\end{itemize}
    
\paragraph{Image Captioning}
We evaluate our method on COCO~\citep{lin2014microsoft} datasets.
% and NoCaps~\citep{nocaps} datasets.
\begin{itemize}
    \item \textbf{COCO} takes the same data split as the image-text retrieval task. 
    % \item \textbf{NoCaps} is used to evaluate models’ capability of describing novel objects which are not seen in the training corpus. This benchmark consists of a validation and test set comprised of 4.5K and 10.6K images.
    % cite refer to oscar

\end{itemize}

\paragraph{Natural Language Understanding}
To verify the natural language understanding ability of our \modelname, we select 6 language understanding datasets from GLUE~\citep{Wang2018GLUEAM} benchmark, including both single-sentence classification tasks and sentence-pair classification tasks.

\begin{itemize}
    \item \textbf{SST-2} The Stanford Sentiment Treebank consists of sentences from movie reviews and human-annotated sentiment. The task is to predict the sentiment of a given sentence. 
    \item \textbf{RTE} The Recognizing Textual Entailment dataset comes from a series of annual textual entailment challenges. 
    \item \textbf{MRPC} The Microsoft Research Paraphrase Corpus consists of a corpus of sentence pairs collected from online news sources, with human annotations for whether the sentences in the pair are semantically equivalent.
    \item \textbf{QQP} The Quora Question Pairs dataset is a collection of question pairs from the community question-answering website Quora. The task is to predict whether a pair of questions are semantically equivalent.
    \item \textbf{MNLI} The Multi-Genre Natural Language Inference Corpus consists of sentence pairs (premise, hypothesis) with textual entailment annotations. The task is to predict the entailment between the premise and the hypothesis.
    \item \textbf{QNLI} The Stanford Question Answering Dataset is a question-answering dataset, where one of the sentences in the paragraph (drawn from Wikipedia) contains the answer to the corresponding question (written by an annotator). The task is to determine whether the context sentence contains the answer to the question.

\end{itemize}

\paragraph{Natural Language Generation}
We use Gigaword dataset~\citep{Rush2015ANA} for text summarization task to verify the natural language generation ability of our \modelname.

\begin{itemize}
    \item \textbf{Gigaword} Headline-generation on a corpus of article pairs from Gigaword consisting of around 4 million articles. It contrains 3803957, 189651 and 1951 samples for training/validation/testing.
\end{itemize}

\paragraph{Video Action Recognition}
We adopt three popular benchmarks Kinetics 400/600/700 dataset~\citep{Kay2017Kinetics} to evaluate our model.

The videos in these three benchmarks are collected from YouTube. Each video clip lasts around 10 seconds and is labeled with a single action class. The videos include human-object interactions such as playing instruments, as well as human-human interactions such as shaking hands and hugging. 

\begin{itemize}
    \item \textbf{Kinetics 400} consists of 240K training videos and 20K validation videos that span 400 human action categories.
    \item \textbf{Kinetics 600} consists of 392K training videos and 30K validation videos spanning 600 action categories.
    \item \textbf{Kinetics 700} consists of 545K training videos and 35K validation videos spanning 700 action categories.
\end{itemize}

\paragraph{Image Classification}
We evaluate performance of \modelname in terms of image classification on ImageNet-1K~\citep{imagenet}.

\begin{itemize}
    \item \textbf{ImageNet-1K} contains 1.28M training images and 50K validation images from 1,000 classes.
\end{itemize}

% \section{Compaison Methods}
% \begin{itemize}
% % text-to-video retrieval
% \item \textbf{Frozen}
% \end{itemize}

% You can have as much text here as you want. The main body must be at most 88 pages long.
% For the final version, one more page can be added.
% If you want, you can use an appendix like this one, even using the one-column format.
%%%%%%%%%%%%%%%%%%%%%%%%%%%%%%%%%%%%%%%%%%%%%%%%%%%%%%%%%%%%%%%%%%%%%%%%%%%%%%%
%%%%%%%%%%%%%%%%%%%%%%%%%%%%%%%%%%%%%%%%%%%%%%%%%%%%%%%%%%%%%%%%%%%%%%%%%%%%%%%

\end{document}